\pdfoutput=1

\documentclass[11pt]{article}

\usepackage[final]{acl}

\def\shownotes{1}

\usepackage{microtype}
\usepackage{graphicx}
\usepackage{caption,subcaption}
\usepackage{booktabs} 
\usepackage{hyperref}
\usepackage{makecell}

\usepackage{textcomp}
\usepackage{multirow}
\usepackage{amsmath,amsthm}

\theoremstyle{definition}

\usepackage{macro}
\usepackage{balance}

\usepackage{algorithm,algorithmicx,algpseudocode}

\usepackage{times}
\usepackage{latexsym}
\usepackage[T1]{fontenc}

\usepackage[utf8]{inputenc}

\usepackage{microtype}

\usepackage{inconsolata}

\usepackage{graphicx}
\makeatletter
\def\thanksnosymbol#1{\protected@xdef\@thanks{\@thanks
        \protect\footnotetext{#1}}}
\makeatother

\title{Efficient Ensemble for Fine-tuning Language Models on Multiple Datasets}

\author{Dongyue Li\textsuperscript{\dag}\thanksnosymbol{Email correspondence to {\{li.dongyu, zhang.zini, ho.zhang\}@northeastern.edu} and {wangluxy@umich.edu}.} \ \ \ \ Ziniu Zhang\textsuperscript{\dag} \ \ \ \ 
Lu Wang\textsuperscript{\ddag} \ \ \ \ Hongyang R. Zhang\textsuperscript{\dag} \\
        \textsuperscript{\dag}Northeastern University, Boston, MA\\
        \textsuperscript{\ddag}University of Michigan, Ann Arbor, MI
}

\begin{document}
\maketitle

\begin{abstract}
This paper develops an ensemble method for fine-tuning a language model to multiple datasets. Existing methods, such as quantized LoRA (QLoRA), are efficient when adapting to a single dataset. When training on multiple datasets of different tasks, a common setup in practice, it remains unclear how to design an efficient adaptation for fine-tuning language models. We propose to use an ensemble of multiple smaller adapters instead of a single adapter per task. We design an efficient algorithm that partitions $n$ datasets into $m$ groups, where $m$ is typically much smaller than $n$ in practice, and train one adapter for each group before taking a weighted combination to form the ensemble. The algorithm leverages a first-order approximation property of low-rank adaptation to quickly obtain the fine-tuning performances of dataset combinations since methods like LoRA stay close to the base model. Hence, we use the gradients of the base model to estimate its behavior during fine-tuning. Empirically, this approximation holds with less than $1\%$ error on models with up to $34$ billion parameters, leading to an estimation of true fine-tuning performances under $5\%$ error while speeding up computation compared to base fine-tuning by $105$ times. When applied to fine-tune Llama and GPT models on ten text classification tasks, our approach provides up to $10\%$ higher average test accuracy over QLoRA, with only $9\%$ more FLOPs. On a Llama model with $34$ billion parameters, an ensemble of QLoRA increases test accuracy by $3\%$ compared to QLoRA, with only $8\%$ more FLOPs. %
\end{abstract}

\section{Introduction}

Parameter-efficient fine-tuning has emerged as an efficient approach to adapting a large language model (LLM) to a downstream application.
In practice, one often runs into a situation where the evaluation criterion involves a mix of many datasets or tasks.
Thus, it would be desirable to have a fine-tuning method that can deliver robust performance in the presence of multiple training datasets.
Directly applying a base fine-tuning method such as low-rank adaptation \cite{hu2021lora} or adapter-tuning \cite{houlsby2019parameter} can lead to negative interference across tasks, as the optimal adapter weights for different datasets can conflict with each other \cite{wuunderstanding,yang2020precise}.
In this paper, we study the design of ensemble methods for parameter-efficient fine-tuning of language models in the presence of multiple datasets.

\begin{figure*}[t!]
    \centering
    \includegraphics[width=0.99\textwidth]{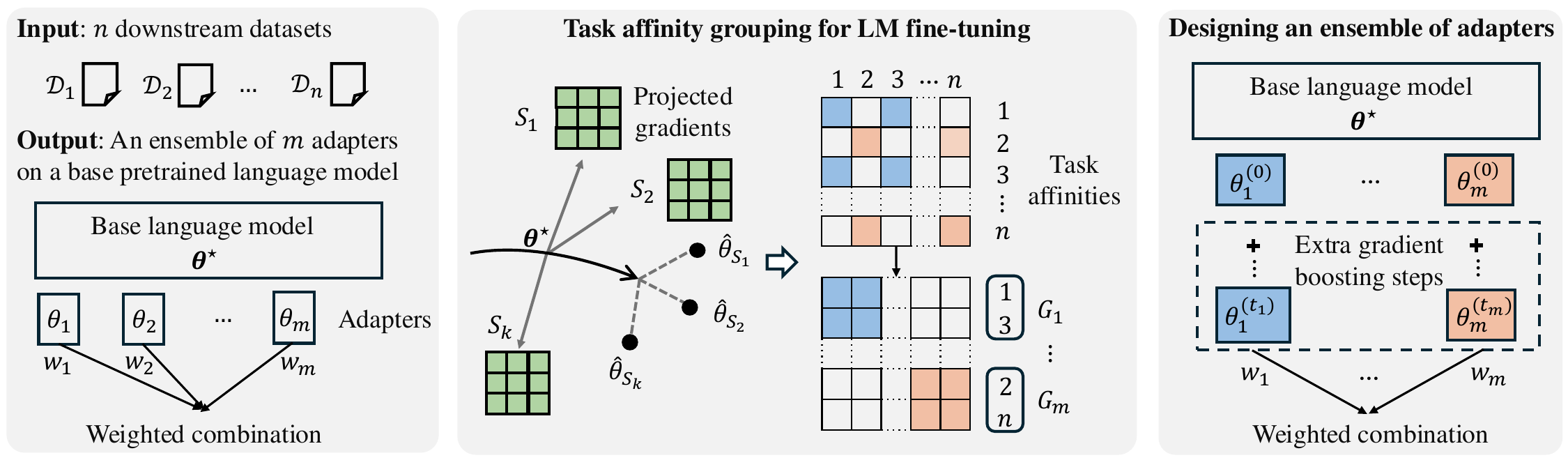}
    \caption{ \textbf{Left}: We propose an ensemble method for fine-tuning language models on multiple datasets. Given $n$ datasets and a base adaptation method such as LoRA, we design an ensemble of adapters that applies weighted averaging to their outputs, with minimal computation and memory overhead to the base method. 
    \textbf{Middle:} We partition $n$ datasets into $m$ groups based on the task affinities scores.
    Our method first estimates fine-tuning performances on multiple dataset combinations, $S_1, S_2, \dots, S_k$, by evaluating the gradients of the adapter weights at $\theta^{\star}$. For each subset $S_i$, we estimate the fine-tuned adapter weights $\hat\theta_{S_i}$ by solving a regression problem, using the projected gradients within $S_i$ as features.
    This leads to an $n$ by $n$ task affinity matrix $T$, where $T_{i,j}$ is the affinity score computed from the estimates (see equation \eqref{eq_affinity_score}). We then partition the $n$ datasets into $m$ groups with a clustering algorithm applied to $T$. In practice, $m$ is usually much smaller than $n$.
    \textbf{Right:} We design an adapter ensemble by fine-tuning one adapter per group and further refining it with a few gradient-boosting steps.
    This overall procedure incurs little computational overhead, as we will describe in Table \ref{table_compare}.}\label{fig_pipeline}
\end{figure*}

Besides being a common problem for evaluating LLMs \cite{hendrycks2020measuring}, other cases where the above problem arises include fine-tuning transformer models on human behavior data and cognitive psychology benchmarks \cite{codacogbench}, as well as reinforcement learning \cite{sodhani2021multi}, and embodied reasoning such as task and motion planning.

One natural approach to the above problem would be to first pre-train a shared adapter on all the datasets and then fine-tune this adapter specifically on each dataset, akin to the paradigm of multitask pretraining followed by supervised fine-tuning (namely MTL-FT in short) \cite{wang2018glue,liu2019multi}.
However, this would increase the computational overhead by a factor of two (one for pretraining and the other for task-specific fine-tuning), which would be costly when the dataset sizes are large.
Directly using a parameter-efficient fine-tuning method such as LoRA \cite{hu2021lora} or QLoRA \cite{dettmers2024qlora} can lead to negative transfer as the weights between the low-rank factors trained on one dataset can differ from another (See Section \ref{sec_prelim} for our experiments confirming this point).

A key technical challenge to ensure robust generalization across $n$ datasets is to understand the relationship between the $n$ datasets when they are trained on top of an LLM.
In this vein, one might also use influence functions \cite{koh2017understanding}, which would require scaling up influence computation to subsets of dataset combinations.
Recent developments in data modeling \cite{ilyas2022datamodels} show that by randomly sampling subsets of data, one can estimate the influence of adding or removing one sample from the dataset upon the trained model efficiently \cite{park2023trak,li2023identification}.
Inspired by this line of work, we propose an efficient estimation of model fine-tuning performances \emph{{without performing any fine-tuning}}.
We begin by noting that for fine-tuning transformer models such as Llama and GPT, the (low-rank) adapter weights typically stay very close to the base model ($\approx 0.2\%$ evaluated on a Llama model with $34$ billion parameters). %
Thus, one can expand the loss using first-order Taylor's expansion around the initialization and then use the gradients measured at the base model to approximate model fine-tuning behavior.
Empirically, we find that this approximation incurs less than $5\%$ error for a variety of fine-tuning methods, including LoRA, adapter-tuning, quantized LoRA, and quantized adapter.

Provided with efficient estimation of fine-tuning performances on dataset combinations, we then design an ensemble of low-rank adapters.
At a high level, given $n$ datasets, our algorithm first partitions the $n$ datasets into $m$ similar groups (where $m$ is typically much less than $n$).
Concretely, we achieve this based on a task affinity matrix that we estimated using the above first-order approximation property.
Then, we fine-tune one adapter in each group, leading to $m$ adapters in total.
We further refine the adapters by applying a few gradient-boosting steps to reduce the loss of groups with the highest training losses.
Importantly, except for computing the gradients at the base model, which we use during the estimation, the rest of this procedure can be done entirely on CPUs.
Additionally, the memory overhead is now roughly $m$ times the size of each adapter.
See Figure \ref{fig_pipeline} for an illustration of our approach.

We extensively validate our approach, showing its broad applicability in boosting a base fine-tuning method with little computational overhead.
First, we show that our estimation method approximates fine-tuning losses within 5\% relative error while reducing computation by $105\times$ compared to actual fine-tuning. 
Second, for fine-tuning Llama models across ten tasks from SuperGLUE, our approach improves QLoRA’s average test accuracy by $\textbf{10}\%$, incurring only $9\%$ additional computation and 9 GB additional memory. Compared to two commonly used methods \cite{liu2019multi,fifty2021efficiently}, our method achieves similar accuracy with $45\%$ less computation and memory.
Further, we scale our approach to a federated learning setting with $500$ datasets, observing qualitatively similar results.
Finally, we analyze the generalization of low-rank adapters by measuring their empirical generalization errors and their sharpness. We find that small-rank LoRA adapters consistently yield the lowest generalization errors across tasks. An ensemble of low-rank adapters further reduces generalization errors compared to one adapter of the same size. 
We provide the code implementation of this work at \href{https://github.com/VirtuosoResearch/EnsembleLoRA}{github.com/VirtuosoResearch/EnsembleLoRA}.

\begin{table}[t!]
\centering
\resizebox{\columnwidth}{!}
{
\begin{tabular}{@{}l c c c c @{}}
\toprule
\textbf{Approach} &  \makecell{\textbf{Runtime}} & \textbf{Memory} \\ \midrule 
Base fine-tuning & $T$ & $A$&\\
Pre-train then fine-tune & $\approx 2 T$ & $n A$\\
Our ensemble method  & ${T + G}$ & ${M A}$\\
\bottomrule
\end{tabular}
}
\caption{A summary of the runtime and memory usage by our approach. $T$ is the runtime of the base fine-tuning method; $G$ is the cost of evaluating the gradients once for all the samples at the base model. In practice, $T$ is usually much higher than $G$. $A$ is the memory used by one adapter, including activation memory. $M$ is the number of adapters in the ensemble.}\label{table_compare}
\end{table}

\section{Methods}\label{sec_approach}

We now describe our approach to designing an ensemble of adapters that applies to any base adaptation method.
Our method quickly partitions $n$ datasets into $m$ groups.
Then, we train one adapter for each group, with the option of adding a few additional adapters via gradient boosting.
Finally, we take a weighted combination of all the adapters as the ensemble output.

\subsection{Preliminaries}\label{sec_prelim}

We present a case study by fine-tuning Llama-3-8B on SuperGLUE \cite{wang2019superglue}.
We collect ten datasets, covering five categories of sentence completion (COPA, H-SWAG, and Story Cloze datasets), natural language inference (CB and RTE), coreference resolution (WSC and Winogrande), question answering (BoolQ and MultiRC), and word sense disambiguation (WiC).
We evaluate LoRA, adapter, QLoRA \cite{dettmers2024qlora}, and quantized adapter (QAdapter). In particular, QLoRA applies $4$-bit quantization to the base model, with LoRA added to the quantized base model.
We evaluate memory usage with a batch size of $4$ with a sequence length of $512$.%

First, we compare (i) fine-tuning one model on each single task and (ii) fine-tuning one model on each pairwise combination of one task with the other nine tasks.
When using full fine-tuning, we observe that for $20$ (out of $45$ pairs), (ii) performs worse than (i). For LoRA and adapter, this increases to $27$ and $24$. For QLoRA and Qadapters, this further increases to $33$ and $35$, respectively.

Second, we show that an ensemble of adapters can improve task performance but can increase memory cost.
For each method, we train one adapter on all tasks and fine-tune it on each individual task initialized from the pretrained adapter.
Then, we apply a weighted combination to the outputs of all adapters to form the ensemble. 
With QLoRA, the ensemble boosts accuracy by 10.8\% over a single QLoRA adapter. With QAdapter, it improves accuracy by 9.5\%.
However, this uses 4 times more memory since it combines the outputs from $10$ adapters simultaneously.

The observations indicate that a single adapter underperforms full fine-tuning due to interference between different datasets. If we use a separate adapter for each dataset, the overhead is proportional to training $n$ adapters.
Thus, the key challenge of adapting an LLM to multiple datasets is to reduce this overhead in a computationally efficient way.
We propose to achieve this by grouping similar datasets so that we can share one adapter for each group of datasets.

\subsection{Task affinity grouping for LM fine-tuning}\label{sec_estimation}

The first step of our approach is to partition the $n$ given datasets into $m$ disjoint subsets based on their affinities trained on top of a language model. %
We estimate affinity scores between dataset $i$ and $j$ for every $i, j \in \set{1, 2, \dots, n}$ as follows. Let $S$ be a subset of datasets containing $i$ and $j$. Denote $f_i(S)$ as the fine-tuning performance on dataset $i$ (e.g., validation loss) using all the datasets in $S$. Let the task affinity score between $i$ and $j$ be:
\begin{align}\label{eq_affinity_score}
    T_{i, j} = \frac{1}{n_{i, j}} \sum_{S_k: i\in S_k, j \in S_k} f_i(S_k),
\end{align}
where $\set{S_k}$ are subsets sampled from $\set{1, \dots, n}$ and $n_{i,j}$ is the number of random subsets that contain both $i,j$.
This score is analogous to the feature importance scores in random forests.

Computing ${T_{i, j}}$ requires fine-tuning the model on multiple subset combinations. Instead, we design an algorithm to estimate $f_i(S)$, which can run on CPUs and deliver estimation results for fine-tuning billion-parameter models in seconds.
The key idea is based on the observation that fine-tuned adapters stay close to the base model. Thus, applying a first-order Taylor's expansion to the base model loss yields a negligible error, which allows us to approximate fine-tuning performance using gradients calculated on the base model.

We now measure the relative distance from the fine-tuned model weights to the base model, i.e., $\frac{\normFro{X-\theta^\star}}{\normFro{\theta_{\text{Full}}^\star}}$, where $X$ is parameter of the fine-tuned adapter, $\theta^{\star}$ is the initialized parameter of the adapter, and $\theta_{\text{Full}}^{\star}$ is the parameter of the entire pretrained model.
As shown in Table \ref{tab_finetuning_distance}, we find that the relative distance remains less than $0.2\%$ on average, across LoRA, adapter, QLoRA, and QAdapter on random data subsets.

\begin{table}[h!]
\centering
\caption{We report the distance between fine-tuned model weights relative to the base model. The results are averaged over 50 sampled task subsets of size 3.} \label{tab_finetuning_distance}
\resizebox{0.475\textwidth}{!}
{
\begin{tabular}{lcccccccc}
\toprule
 & Llama-3-1B & Llama-3-3B & Llama-3-8B \\ \midrule
LoRA & $ 0.16_{\pm 0.04}$\% & $ 0.14_{\pm  0.02}$\% &  $0.12_{\pm 0.02}$ \% \\
QLoRA & $ 0.18_{\pm 0.02}$\% & $ 0.16_{\pm  0.03}$\% &  $0.11_{\pm 0.02}$ \%  \\
Adapter & $ 0.09_{\pm 0.03}$\% & $ 0.05_{\pm  0.01}$\% &  $0.08_{\pm 0.02}$ \% \\
QAdapter & $ 0.11_{\pm 0.03}$\% & $ 0.08_{\pm  0.03}$\% &  $0.07_{\pm 0.01}$ \%  \\
\bottomrule
\end{tabular}
}
\end{table}

\begin{figure}[t!]
    \begin{subfigure}[b]{0.235\textwidth}
    \centering
    \includegraphics[width=0.98\textwidth]{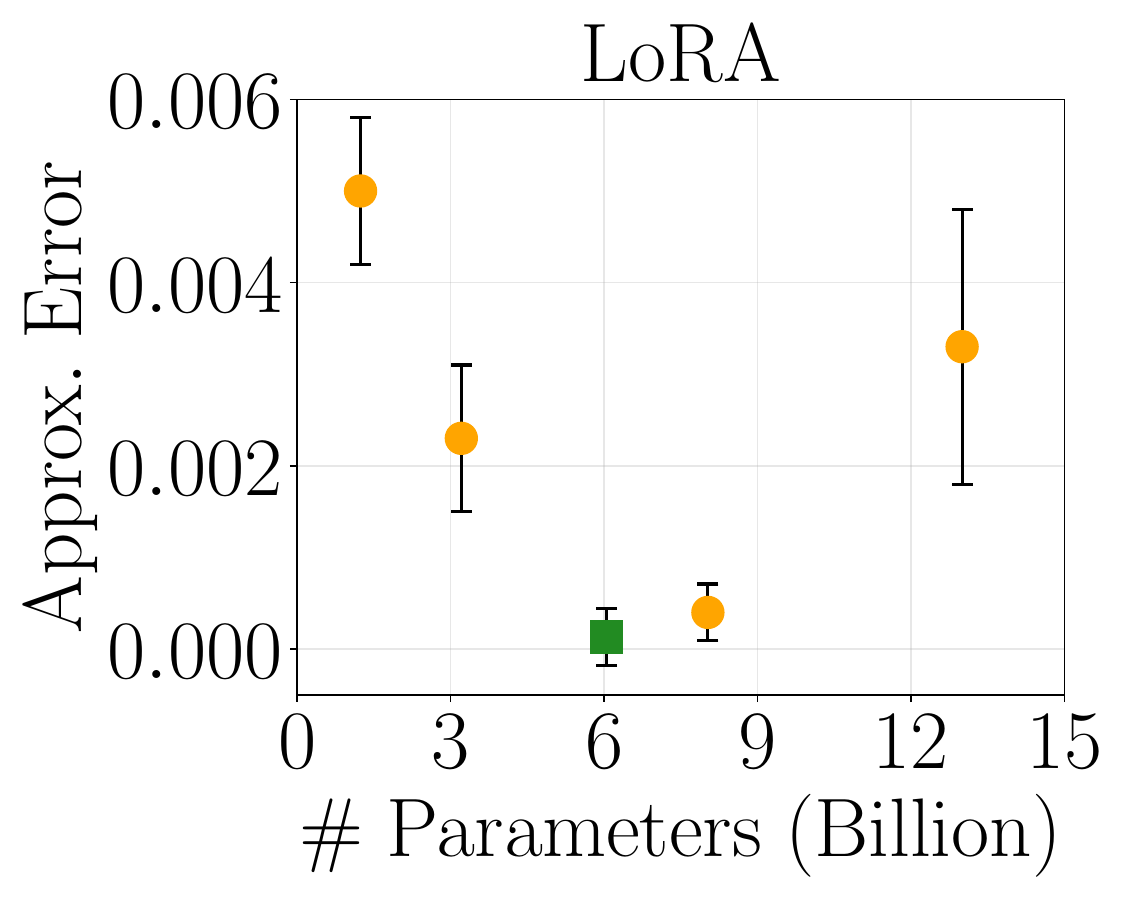}
    \end{subfigure}
    \begin{subfigure}[b]{0.235\textwidth}
    \centering    
    \includegraphics[width=0.98\textwidth]{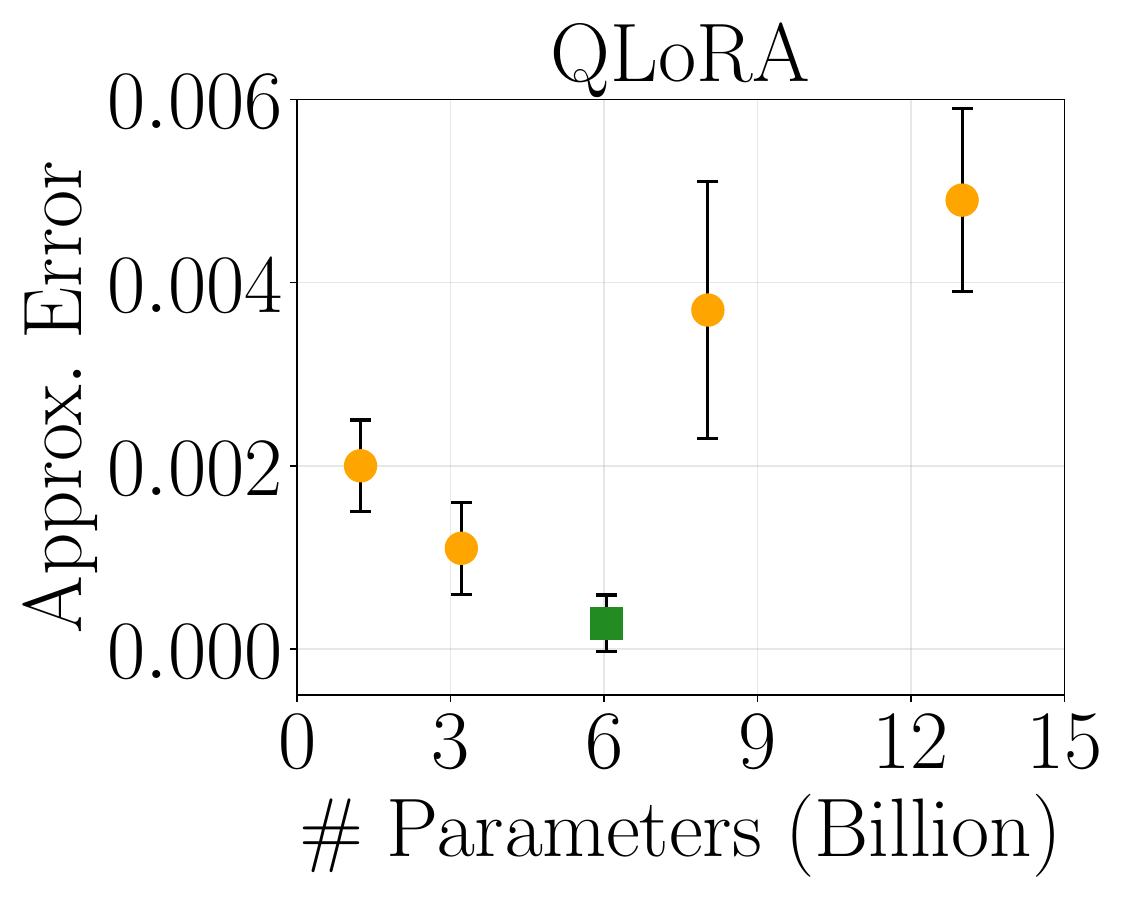}
    \end{subfigure}
    \begin{subfigure}[b]{0.235\textwidth}
    \centering    
    \includegraphics[width=0.98\textwidth]{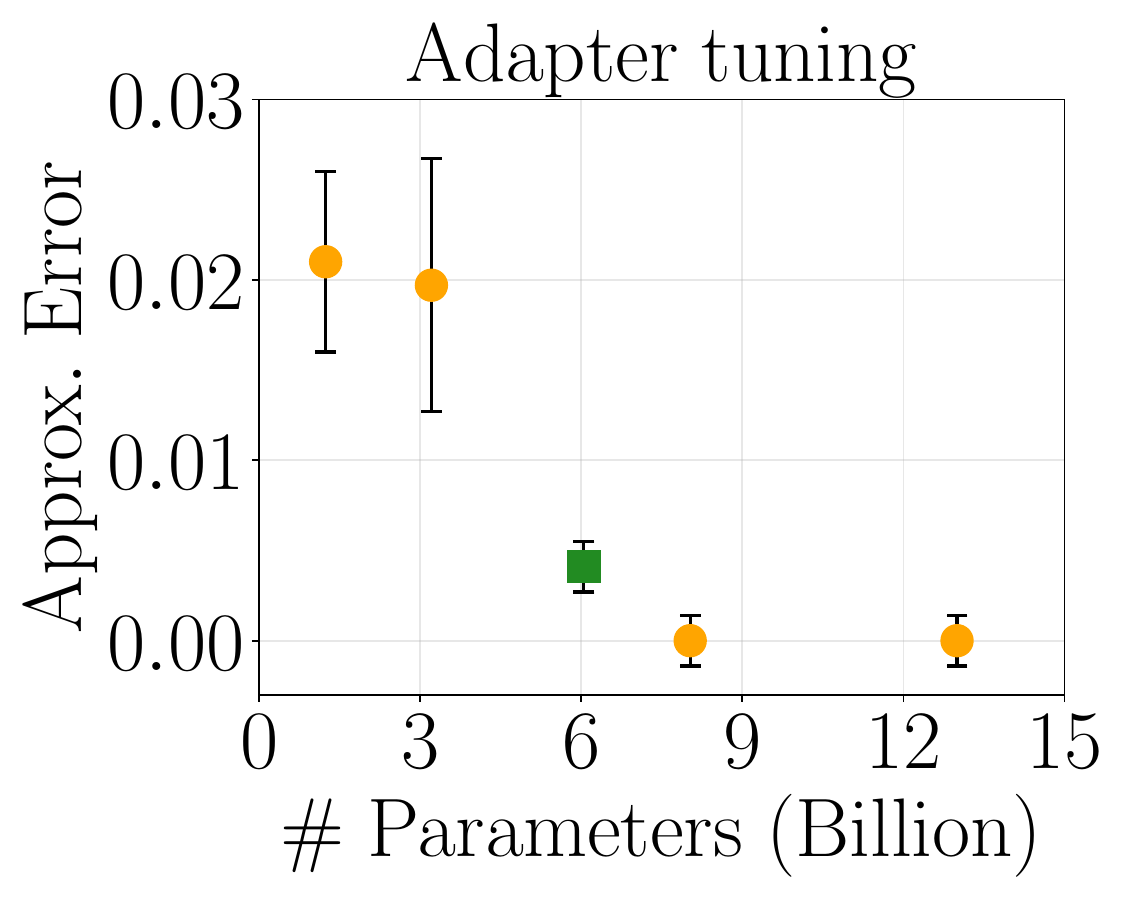}
    \end{subfigure}
    \begin{subfigure}[b]{0.235\textwidth}
    \centering
    \includegraphics[width=0.98\textwidth]{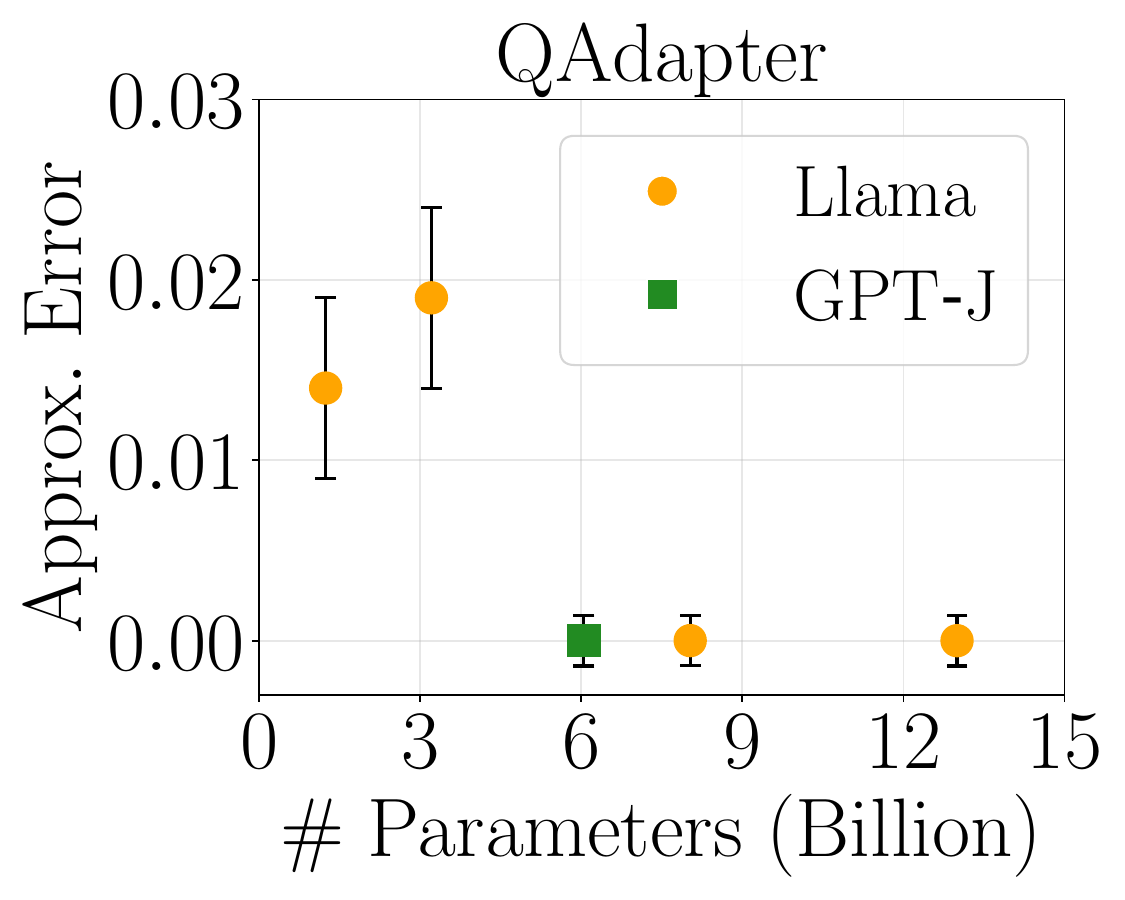}
    \end{subfigure}
    \caption{We report the approximation error for Llama and GPT-J models with up to 34 billion parameters for LoRA, QLoRA, adapter, and QAdapter. %
    We report the average and the standard deviation based on the results from $50$ randomly sampled task subsets of size $3$.}\label{fig_approximation_pretrained_models}
\end{figure}

\noindent\textbf{A gradient-based estimation algorithm:}
Let the output of a neural network be given by $h_X(s, y)$, where $s$ is an input such as a sentence and $y$ is the ground-truth label. 
Suppose that $X\in \real^p$ is the trainable parameter in an adapter.
The Taylor's expansion of $h_X(s, y)$ centered at $\theta^{\star}$ is given as: 
\begin{align*}%
      h_X(s, y) \approx h_{\theta^{\star}}(s, y) & + [\nabla_X h_{\theta^{\star}}(s, y)]^{\top} (X - \theta^{\star}) \\
      & + \epsilon_s.
\end{align*}
Next, we evaluate the magnitude of $\epsilon_s$ relative to $h_{X}(s, y)$ and report the results in Figure \ref{fig_approximation_pretrained_models}.
We illustrate the error for six language models with up to 34 billion parameters.
We find that for both Llama and GPT-J models, the approximation error is less than $1\%$ in all cases. %
For QLoRA and QAdapter, the approximation error is less than 3\%.

Since the approximation error is small, we next apply it to approximate the model loss.
We will illustrate the case for log loss with binary classification, and the same idea applies to other types of loss functions.
Let $y \in \set{+1, -1}$. Recall the log loss is $\log\left( 1 + \exp\left(-y \cdot h_{X}(s, y) \right) \right)$.
Define $b_s = - y \cdot h_{\theta^{\star}}(s, y)$ and $g_s = \nabla h_{\theta^{\star}}(s, y)$.
Using Taylor's expansion without the higher-order terms, we approximate $h_X(s, y)$ as: 
\begin{align*}\label{eq_estimation_logistic_regression}
   \hat\ell(X) 
   =  \log\left(1 + \exp\left(b - y\cdot g^{\top} (X - \theta^{\star}) \right) \right).
\end{align*}
For a subset $S \subseteq \set{1, 2, \dots, n}$ with training samples denoted as $\cD_S$, we estimate fine-tuned adapter weights by minimizing averaged $\hat\ell(X)$ on $\cD_S$:
\begin{align*}
    \hat{\theta}_S \leftarrow \mathop{\arg\min}_{X \in \real^p} \frac{1}{n_S} \sum_{(s, y) \in \cD_S} \hat \ell(X).%
\end{align*}
Using $\hat \theta_S$, we evaluate the estimated fine-tuning performance $\hat f_i(S)$ for every task $i \in S$. 
Then, we apply this estimation for $k$ random subsets sampled from $[n]$ to compute $T_{i,j}$ for every $i$ and $j$. This forms a ${n\times n}$ task affinity matrix $T$. 

Given $T$, we apply a clustering algorithm to partition $n$ datasets into $m$ groups. The clustering objective maximizes the average density of scores within clusters, which is solved based on semi-definite programming relaxations. Additionally, we use a trace regularization term to determine the number of groups $m$.
See details in Appendix \ref{app_approach}.
Let $G_1, G_2, \dots, G_m$ denote the resulting groups.

In summary, we only need to compute the gradients on all the training samples once at the base model.
For each subset, the estimation solves a regression problem based on the gradients.
A random projection is used to reduce the dimension of the gradients down to a few hundred, which provably preserves the Euclidean geometry between the gradient vectors.\footnote{This can be achieved using the Johnson-Lindenstrauss Lemma. After projecting the dimension of the gradients down to several hundred, solving each logistic regression problem will only take a few seconds on a CPU.}
The task affinity grouping procedure is presented in Algorithm \ref{alg_gradient_estimation}.

\subsection{Designing an ensemble of adapters}

The second step of our approach is to compute an adapter ensemble based on the $m$ groups computed above.
We first fine-tune one adapter on tasks in each group, yielding $m$ adapters denoted as $\theta_1^{(0)}, \theta_2^{(0)}, \ldots, \theta_m^{(0)}$.

\begin{algorithm}[t!]
\raggedright
\caption{Efficient task affinity grouping for language model fine-tuning}\label{alg_gradient_estimation}
\textbf{Input}: $n$ training/validation datasets\\
\textbf{Require:} Pretrained model $h_{\theta^\star}$, number of subsets $k$, and projection dimension $d$\\
\begin{algorithmic}[1]
    \State $\set{S_1, S_2, \ldots, S_k} \leftarrow$ sample a list of subsets from $\set{1,2,\dots,n}$ with a fixed size
    \State $P \leftarrow$ $p$ by $d$ Gaussian random matrix
    \For{$(s, y) \in \cD_{\set{1, 2, \dots, n}}$}
        \State $\tilde{g}_s \leftarrow P^{\top} \nabla h_{\theta^{\star}}(s, y)$ \Comment{\hspace{-0.01in}project the gradient}%
        \State $b_s \leftarrow - y\cdot h_{\theta^{\star}}(s, y)$
    \EndFor
    \For{$S \in \set{S_1, S_2, \dots, S_k}$}
    \State $\hat X_{d} \leftarrow$ solve a regression problem based on $(\tilde g_s, b_s, y)$ for all $(s, y) \in \cD_{S}$
    \State $\hat \theta_{S} \leftarrow \theta^{\star} + P \hat X_d$ \Comment{restore the dimension}
    \State $\hat{f}_i(S) \leftarrow$ evaluate $h_{\hat\theta_S}$ on the validation set for each $i \in S$  %
    \EndFor

    \State $T\leftarrow $ compute an $n$ by $n$ matrix $T$ following equation \eqref{eq_affinity_score} 
    \State $\set{G_1, G_2, \dots, G_{m}} \leftarrow$ apply a clustering algorithm on $T$ to obtain a partition of $n$ tasks 
    \State Return $G_1, G_2, \dots, G_{m}$
\end{algorithmic}
\end{algorithm}

Next, we apply $b$ gradient boosting steps to reduce the loss of groups with high training losses.
At each step $i$, we choose a group $G_{j_i}$ from $G_1, G_2, \dots, G_m$ that has the largest training error, where $j_i$ is the index of the group chosen at step $i$. 
Suppose there are $t_{j_i}$ adapters for group $G_{j_i}$. 
We then fit a new adapter to predict $1 - p_s$ (the negative gradient of log loss), where $p_s$ is the correct class prediction probability of the current model on a sample $(s, y)$, by minimizing the mean squared loss on group $G_{j_i}$: 
\begin{align*}
    \min_X \frac{1}{n_{G_{j_i}}} \sum_{(s, y) \in \cD_{G_{j_i}}} \Big(1 - p_s - h_{X}(s, y)) \Big)^2.
\end{align*}
We approximate the above loss based on the gradient estimation, leading to the following linear regression on group $G_{j_i}$:
\begin{align*}%
    \hat \ell(X) = 
    \Big(1 - p_s - h_{\theta^{\star}}(s, y) - g_s^{\top} (X - \theta^{\star} ) \Big)^2. 
\end{align*}
Let the minimizer of the above averaged over $G_{j_i}$ be denoted as $\theta_{j_i}^{(t_{j_i} + 1)}$. 
After obtaining this new adapter, we add it along with $\theta_{j_i}^{(0)}, \ldots, \theta_{j_i}^{(t_{j_i})}$ to form the ensemble for group $G_i$. The prediction for this group is based on adding all the outputs from the ensemble. In our ablation study of applying the boosting procedure on three groups of ten tasks, we found that the first boosting step alone reduces training error by $18\%$, leading to $0.4\%$ average test accuracy improvement for tasks in one group.

The final ensemble contains a total of $M = m + b$ adapters. We then train the weights to form a weighted combination of the $m$ groups.
Along with the first step, we summarize the complete procedure in Algorithm \ref{alg_boosting}.
\begin{algorithm}[t!]
\caption{\algname~(Ensemble of Low-Rank Adapters for multiple datasets)}\label{alg_boosting}
\raggedright
\textbf{Input}: $n$ training/validation datasets\\
\textbf{Require}: Pretrained model $h_{\theta^\star}$, number of subsets $k$, projection dimension $d$, boosting steps $b$  \\
\begin{algorithmic}[1]
    \State $G_1, G_2, \dots, G_{m} \leftarrow$ apply Algorithm \ref{alg_gradient_estimation} to the $n$ input datasets
    \State $\theta_1^{(0)}, \dots, \theta_{m}^{(0)} \leftarrow$ Fine-tune one adapter for each group in $G_{1}, \dots, G_{m}$
    \For{$i = 1, \dots, b$} \Comment{Boosting steps}
        \State $G_{j_i} \leftarrow$ select the group with the highest training loss, where $j_i$ is the group index
        \State $\theta_{j_i}^{(0)}, \ldots, \theta_{j_i}^{(t_{j_i})} \leftarrow$ get all the $t_{j_i} + 1$ adapters currently in $G_{j_i}$
        \State $\theta_{j_i}^{(t_{j_i} + 1)} \leftarrow$ estimate an adapter to fit the negative gradients of samples in  $\cD_{G_{j_i}}$ %
    \EndFor
    \State $w_1, w_2, \dots, w_{m} \leftarrow$ train the weights for combining outputs of each group of adapters  
    \State Return $\big\{ \theta_i^{(0)}, \dots, \theta_i^{(t_i)}, w_i \big\}_{i=1}^{m}$
\end{algorithmic}
\end{algorithm}

\section{Experiments}\label{sec_exp}

We evaluate our approach to answer the following questions.
How accurate are the gradient-based estimates relative to the actual fine-tuning losses? 
What is the trade-off between performance, computation cost, and memory overhead between a base fine-tuning method and its ensemble?

As a summary of our empirical findings, we show that our approach estimates fine-tuning losses within \textbf{5}\% error while using \textbf{105}$\times$ less computation than actual fine-tuning.
For fine-tuning Llama models on SuperGLUE, \algname{} boosts the accuracy of QLoRA and QAdapter by up to \textbf{10}\%, with 9\% additional computation and 9 GB extra memory usage.
Compared with pretraining followed by fine-tuning, our approach yields comparable performance, reducing computation and memory both by \textbf{45}\%.
Our approach can scale to $500$ datasets and remains comparable to pretraining followed by fine-tuning, reducing computation by \textbf{90}\% and memory by \textbf{91}\%.

\subsection{Experimental setup}
Our approach applies to a wide range of parameter-efficient fine-tuning methods. For a representative evaluation, we focus on the results of using QLoRA, while the results of other methods are quantitatively similar and are deferred to Appendix \ref{sec_exp}. 
We fine-tune language models on ten NLP tasks from SuperGLUE. All the tasks are evaluated as classification tasks. The test accuracy is computed with the provided development set. We then split $10\%$ of the training set as the validation set. The statistics of the datasets are summarized in Table~\ref{tab_full_results_lora}.
We use Llama-3.1-8B and CodeLlama-34B-Instruct as the base model. With a base fine-tuning protocol, we compare our approach against:
Base fine-tuning, which trains a single adapter on the combined set of datasets; 
pretraining followed by fine-tuning (MTL-FT) \cite{liu2019multi}, which performs multitask pretraining and then fine-tunes task-specific adapters from the pretrained adapter;
task affinity grouping (TAG) \cite{fifty2021efficiently}. %

For each baseline, we assess the test accuracy averaged over all tasks.
The error rate is defined as one minus the average test accuracy.
We measure the computational cost as the total number of floating point operations (FLOPs) and the memory cost as the GPU during the inference phase of the model, tested on two Nvidia A6000 GPUs.
In our approach, we sample $k=200$ task subsets of size 3. We vary the number of groups $m$ between $2$ to $6$ and the number of $b$ between 1 to 4.
We provide a complete list of other training parameters used by our approach in Appendix \ref{app_exp}.

\subsection{Estimation results}

We first evaluate the accuracy of Algorithm \ref{alg_gradient_estimation} in estimating fine-tuning losses. Since the first-order approximation incurs a negligible error in model outputs, we show that the estimated fine-tuning performance $\hat{f}_i(S)$ closely matches the actual performance $f_i(S)$. Using test accuracy as the metric, we measure the relative error between $\hat{f}_i(S)$ and $f_i(S)$ over 50 randomly sampled subsets of size 3. For the computation cost, we compare the number of FLOPs between actual full fine-tuning and that required in our estimation procedure.
We experiment with Llama-3-1B, 3B, and 8B models, and vary the projected gradient dimension $d$ between 200, 400, and 800.

\begin{table}[t!]
\centering
\caption{We evaluate the error between estimated and true fine-tuning performances computed on 50 random task subsets. We report the speedup rate between the number of FLOPs used in our approach to full fine-tuning. $d$ is the projection dimension in Algorithm \ref{alg_gradient_estimation}.}\label{tab_approximation_of_finetune_losses}
\resizebox{0.48\textwidth}{!}
{
\begin{tabular}{lccc|ccccc}
\toprule
$d$ & Llama-3-1B & Llama-3-3B & Llama-3-8B & Speedup \\
\midrule
200 & 8.2\% & 8.1\% & 7.0\% & 105$\times$ \\
400 & 4.7\% & 4.8\% & 4.3\% & 105$\times$ \\
800 & 4.6\% & 4.4\% & 4.2\% & 105$\times$ \\
\bottomrule
\end{tabular}
}
\end{table}

\begin{figure}[t!]
    \centering
    \begin{subfigure}[b]{0.48\textwidth}
    \begin{subfigure}[b]{0.48\textwidth}
    \centering    
    \includegraphics[width=0.995\textwidth]{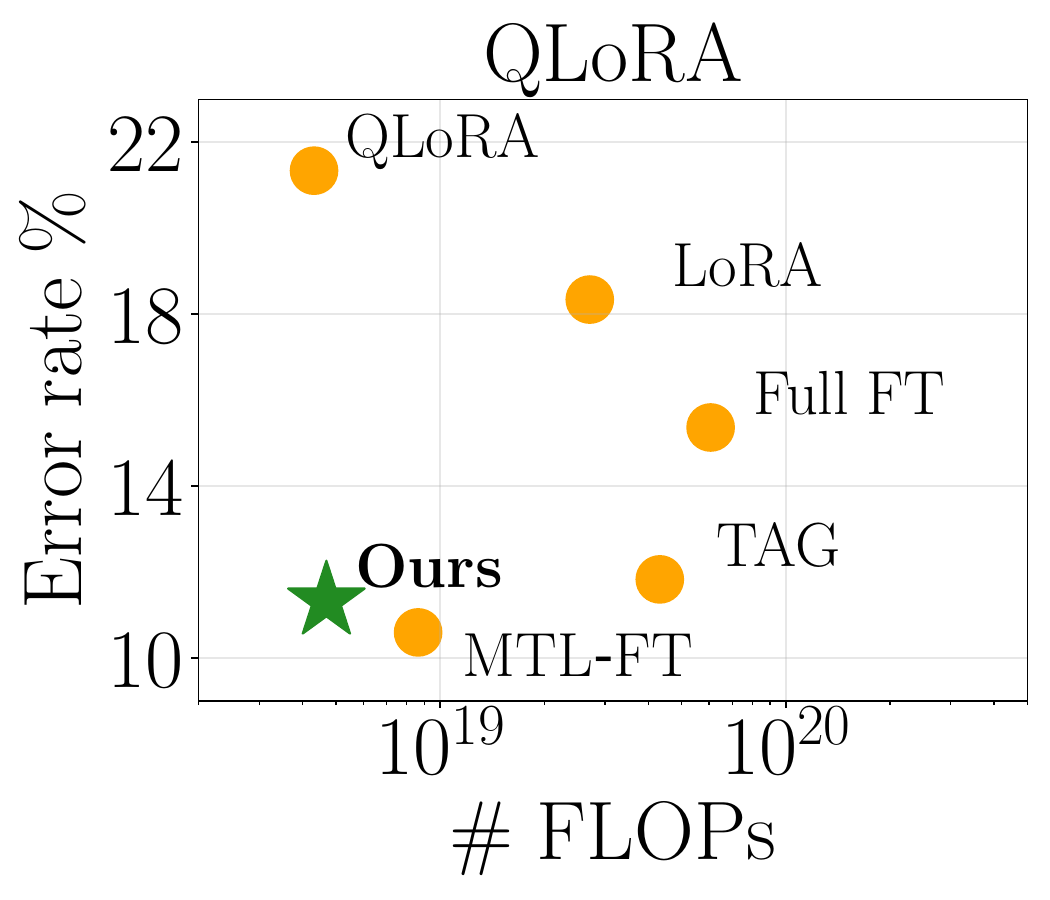}
    \end{subfigure}\hfill
    \begin{subfigure}[b]{0.48\textwidth}
    \centering    
    \includegraphics[width=0.995\textwidth]{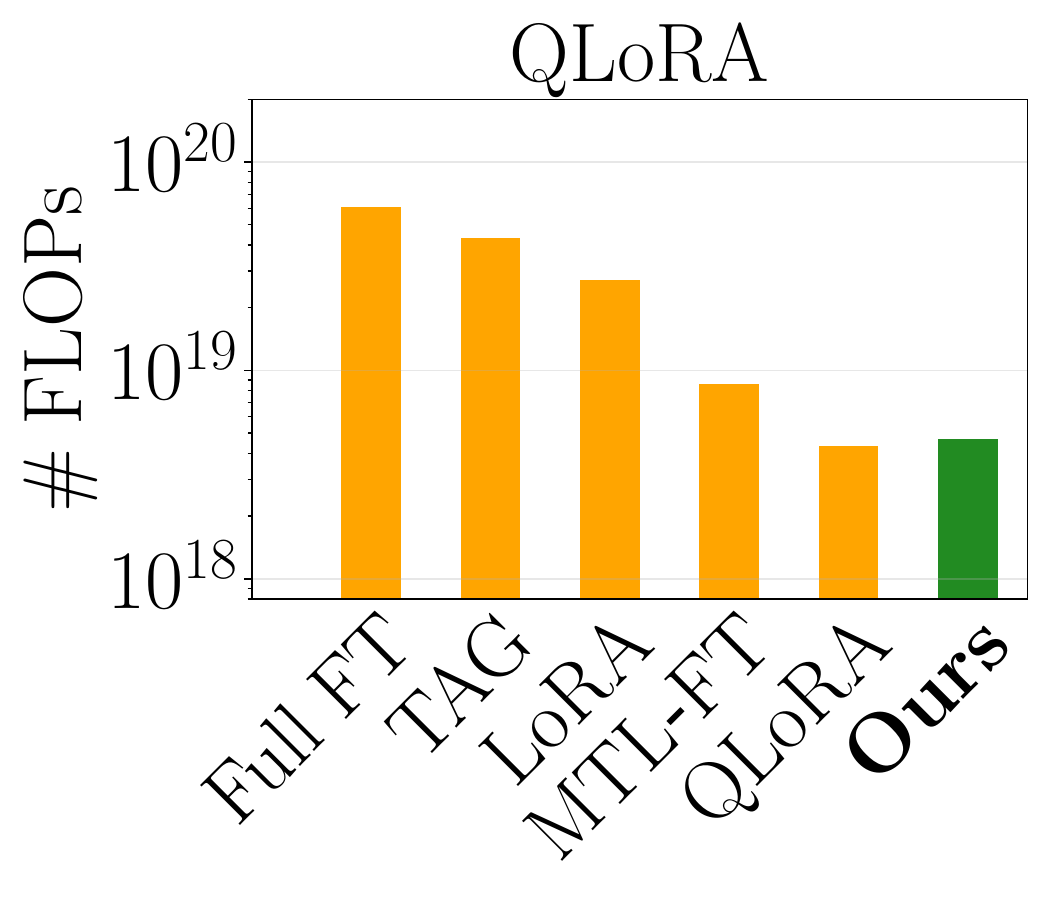}
    \end{subfigure}
    \end{subfigure}
    \begin{subfigure}[b]{0.48\textwidth}
    \begin{subfigure}[b]{0.48\textwidth}
    \centering    
    \includegraphics[width=0.995\textwidth]{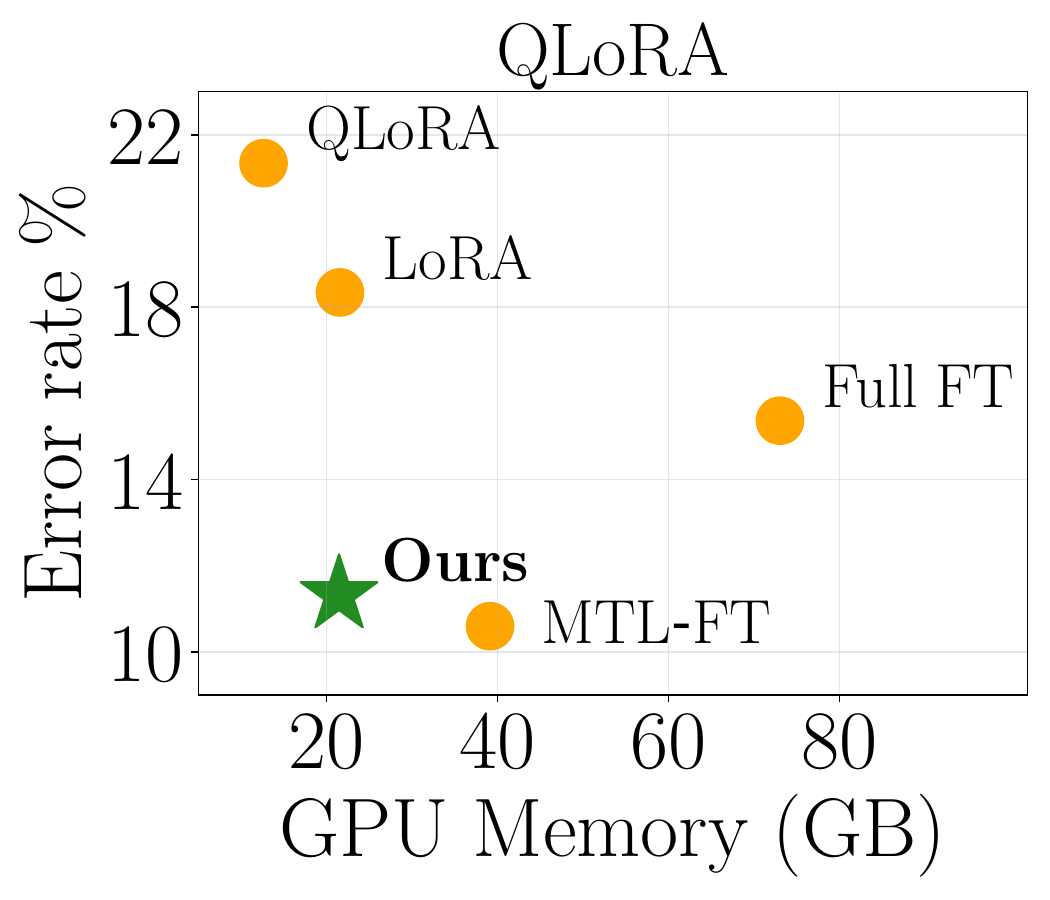}
    \end{subfigure}\hfill
    \begin{subfigure}[b]{0.48\textwidth}
    \centering    
    \includegraphics[width=0.995\textwidth]{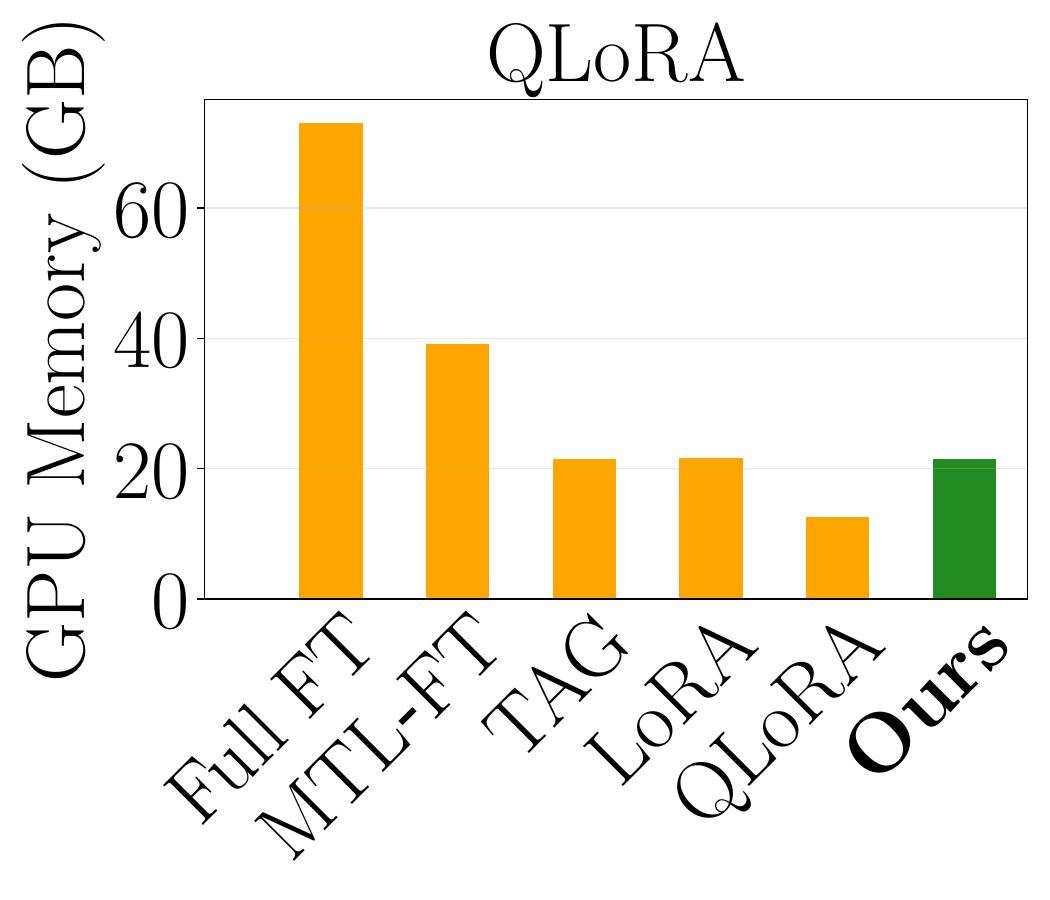}
    \end{subfigure}
    \end{subfigure}
    \caption{We compare error rate (one minus accuracy), computation cost, and memory usage across our approach and baselines when fine-tuning Llama-3-8B on ten NLP tasks. MTL-FT refers to first fine-tuning a shared LoRA on all the datasets, and then fine-tuning the low-rank adapter on each dataset, while Full FT refers to full fine-tuning of the entire model.
    Our approach boosts the test accuracy of QLoRA by $10\%$ on average, only incurring $8\%$ additional computation and 9 GB more memory. It performs on par with the best baseline with $45\%$ less FLOPs.}\label{fig_tradeoff_exp}
\end{figure}

\begin{table*}[t!]
\centering
\caption{We report the test accuracy (\%) of our method, as compared with baselines. We also compute the average test accuracy across ten NLP datasets, along with the number of FLOPs and memory usage for fine-tuning Llama-3-8B using QLoRA. We run each experiment with three random seeds and report the standard deviations.} \label{tab_full_results_lora}
\resizebox{\textwidth}{!}
{
\begin{tabular}{lccccccc|ccc}
\toprule
& BoolQ & CB & COPA & H-SWAG & MultiRC & Story Cloze & Winogrande & 
\multirow{5}{*}{\makecell{Average\\Accuracy}} & \multirow{5}{*}{\makecell{Number of\\ FLOPs}} & \multirow{5}{*}{\makecell{GPU \\ Memory}} \\
\# Train & 4,242 & 225 & 360 & 17,957 & 12,259 & 841 & 4,161\\
\# Validation & 471 & 25 & 40 & 1,995 & 1,362 & 188 & 462\\
\# Test & 3,270 & 56 & 100 & 10,042 & 4,848 & 1871 & 1267\\
\# classes & 2 & 3 & 2 & 4 & 2 & 2 & 2 \\
\midrule
Full FT & 87.6$_{\pm0.9}$ & 93.2$_{\pm0.7}$ & 92.0$_{\pm1.0}$ & 93.2$_{\pm0.6}$ & 86.4$_{\pm0.2}$ & 94.2$_{\pm0.5}$ & 75.1$_{\pm0.2}$ & 84.6$_{\pm 0.9}$  & 6.0$\times 10^{19}$ & 73.0GB \\
LoRA  & 84.5$_{\pm0.2}$ & 87.7$_{\pm0.8}$ & 92.0$_{\pm2.0}$ & 93.4$_{\pm0.3}$ & 82.1$_{\pm0.2}$ & 92.9$_{\pm0.3}$ & 60.7$_{\pm0.4}$ & 81.6$_{\pm 0.8}$ & 2.7$\times 10^{19}$ & 21.5GB \\
QLoRA & 82.0$_{\pm0.5}$ & 83.7$_{\pm0.7}$ & 90.0$_{\pm2.0}$ & 92.3$_{\pm0.3}$ & 84.2$_{\pm0.8}$ & 91.5$_{\pm0.7}$ & 55.1$_{\pm0.3}$ &  78.6$_{\pm 1.2}$ & 4.3$\times 10^{18}$ & 12.6GB \\
MTL-FT & 89.9$_{\pm0.2}$ & 100$_{\pm0.0}$ & 95.0$_{\pm0.5}$ & 93.9$_{\pm0.4}$ & 89.6$_{\pm0.1}$ & 98.0$_{\pm0.5}$ & 84.4$_{\pm0.3}$ & 89.4$_{\pm 0.5}$ & 8.6$\times 10^{18}$ & 39.1GB  \\
TAG & 88.5$_{\pm0.4}$ & 100$_{\pm0.0}$ & 94.0$_{\pm0.5}$ & 92.1$_{\pm0.3}$ & 89.1$_{\pm0.7}$ & 96.3$_{\pm0.4}$ & 80.1$_{\pm0.8}$ & 88.1$_{\pm 0.8}$  & 4.3$\times 10^{19}$ & 21.4GB\\ 
Ours & 89.9$_{\pm0.9}$ & 100$_{\pm0.0}$ & 94.0$_{\pm1.0}$ & 93.5$_{\pm0.3}$ & 89.1$_{\pm0.3}$ & 97.1$_{\pm0.1}$ & 82.0$_{\pm0.5}$ &  88.6$_{\pm 0.6}$ & 4.7$\times 10^{18}$ & 21.4GB\\
\bottomrule
\end{tabular}
}
\end{table*}

Table \ref{tab_approximation_of_finetune_losses} presents the results. 
Our approach achieves a relative error within $9\%$, using \textbf{105}$\times$ less computation than full fine-tuning. We observe that increasing $d$ beyond $400$ reduces the relative error to under \textbf{5}\%, so we set $d=400$ in our experiments. Additionally, we observe that our approach can achieve up to $0.6$ correlation scores with the true fine-tuning performances. 

In particular, our approach uses $6.5 \times 10^{16}$, $4.9 \times 10^{16}$, and $3.2 \times 10^{16}$ FLOPS, when using Llama-3-8B, 3B, and 1B models, respectively. Performing actual fine-tuning requires $6.8 \times 10^{18}$, $5.1 \times 10^{18}, 3.4 \times 10^{18}$, respectively. The speed-up ratio is computed between these two sets of results. 
Recall that our approach involves first evaluating the gradients of all training samples at the base model, and then solving logistic regression on the projected gradients for each subset. The gradient evaluation takes over $80\%$ of the computation. The speed-up is calculated by the ratio between the runtime of actual fine-tuning relative to gradient evaluation, which remains consistent across the three models.

\subsection{Experimental results}

We illustrate the comparison of \algname{} with baselines in Figure \ref{fig_tradeoff_exp}. We report the evaluation results in Table \ref{tab_full_results_lora}. 
Compared to QLoRA fine-tuning, our approach improves accuracy by \textbf{10}\% with only 8\% additional computation and 9 GB more memory.
Compared to pretraining followed by fine-tuning and task grouping, it achieves comparable performance while reducing computation and memory both by 45\%. 
Further, our approach improves over full fine-tuning by \textbf{4.0}\%, with only 92\% less computational cost and 52 GB less memory.
Additionally, our approach also improves the base QAdapter fine-tuning method by \textbf{9}\% accuracy, using 8\% additional computation and 10 GB more memory. 
The quantitative results of using QAdapter are reported in Appendix \ref{app_exp}.

We also apply our approach to a CodeLlama-34B model using QLoRA as the base method. We observe an improved accuracy by \textbf{3.0}\%, while incurring only $8\%$ additional computation and 29 GB more memory. 
Compared to pretraining followed by fine-tuning, \algname{} maintains comparable accuracy while reducing computation by 46\% and memory from 66 GB to 28 GB.

\paragraph{Determining the ensemble size.} We find that the test accuracy stabilizes after $m$ reaches $3$, so we report results using three clusters. We also find that a single boosting step suffices to reduce training error by $18\%$, resulting in a $0.4\%$ increase in the average test accuracy across tasks. This leads to an ensemble of $4$ adapters.
In general, we choose $m$ that maximizes the average density of task affinity scores within clusters, as it correlates with test accuracy in practice. Moreover, the boosting step $b$ is determined by the point where the training loss no longer decreases, which can depend on the number of groups. For settings of fewer groups, a single step is often sufficient.
Once task affinity is estimated, selecting these parameters takes only a few seconds, since no fine-tuning is involved.

\paragraph{Extension.}
We apply our approach to federated learning on a dataset with $500$ tasks, where each speaking role in a play represents a distinct task (client). Data is split $80\%$ for training and $20\%$ for testing. We use QLoRA for fine-tuning, Llama-3-1B as the base model, and the Federated Averaging algorithm. We sample 2,000 subsets of size 10, varying the number of task groups $m$ from $10$ to $30$ and boosting steps $b$ from $1$ to $10$.
 
Our approach reduces test loss over QLoRA by over \textbf{9}\%, while only using 8\% additional computation and 15 GB more memory. 
Compared to pretraining followed by fine-tuning, \algname{} gives comparable test losses while reducing computation by \textbf{90}\% and memory by \textbf{91}\%.

\section{Generalization of Low-rank Adaptation}\label{sec_measure_generalization}

\begin{figure*}[t!]
    \begin{subfigure}[b]{0.33\textwidth}
    \centering
    \begin{subfigure}[b]{0.490\textwidth}
    \includegraphics[width=\textwidth]{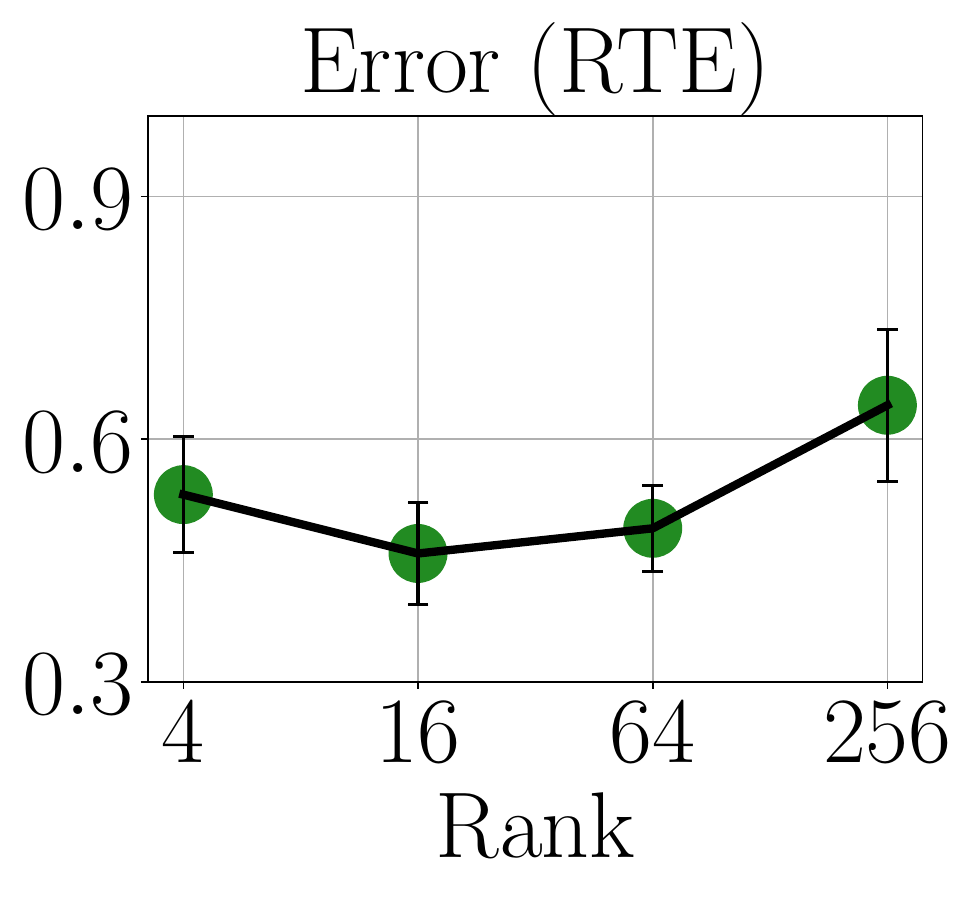}
    \end{subfigure}
    \begin{subfigure}[b]{0.490\textwidth}
    \includegraphics[width=\textwidth]{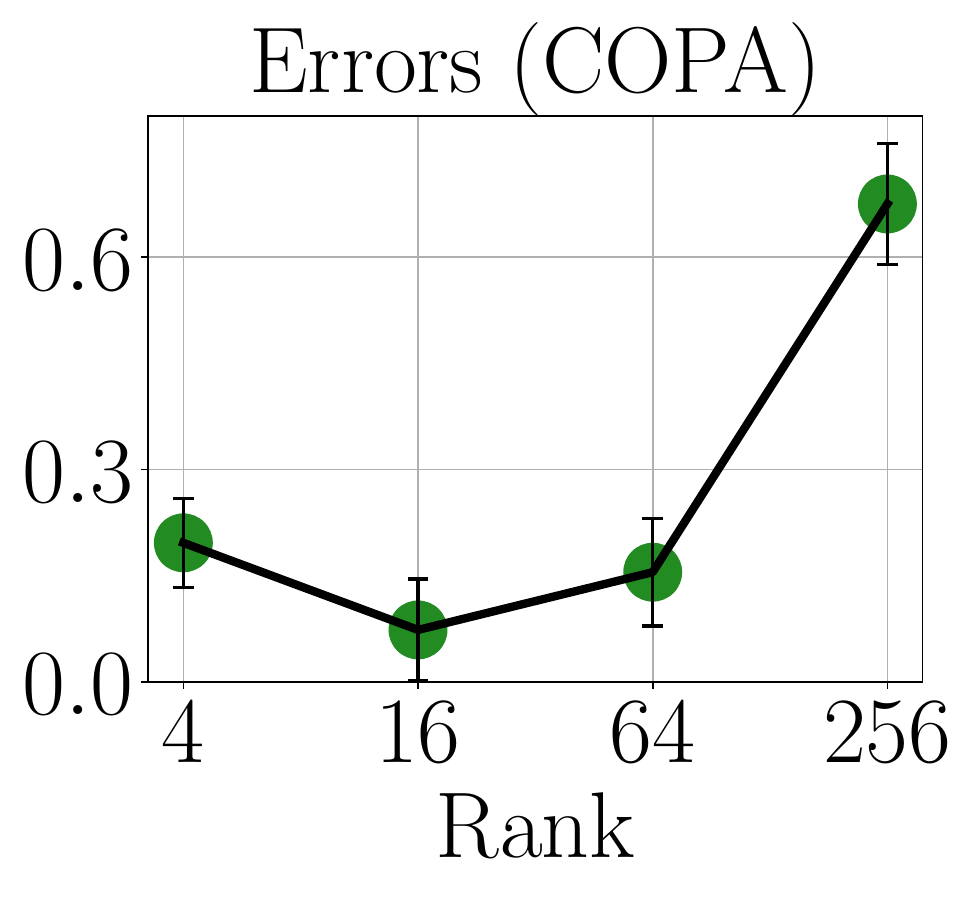}
    \end{subfigure}\vfill
    \begin{subfigure}[b]{0.490\textwidth}
    \includegraphics[width=\textwidth]{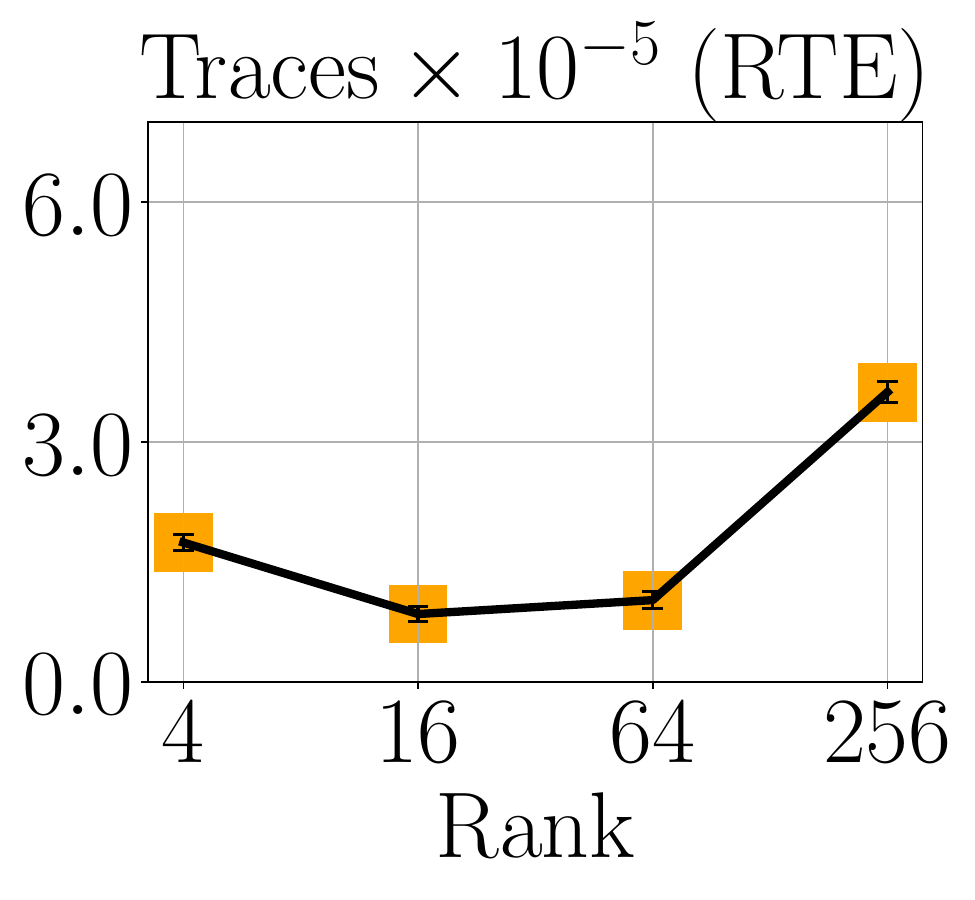}
    \end{subfigure}
    \begin{subfigure}[b]{0.490\textwidth}
    \includegraphics[width=\textwidth]{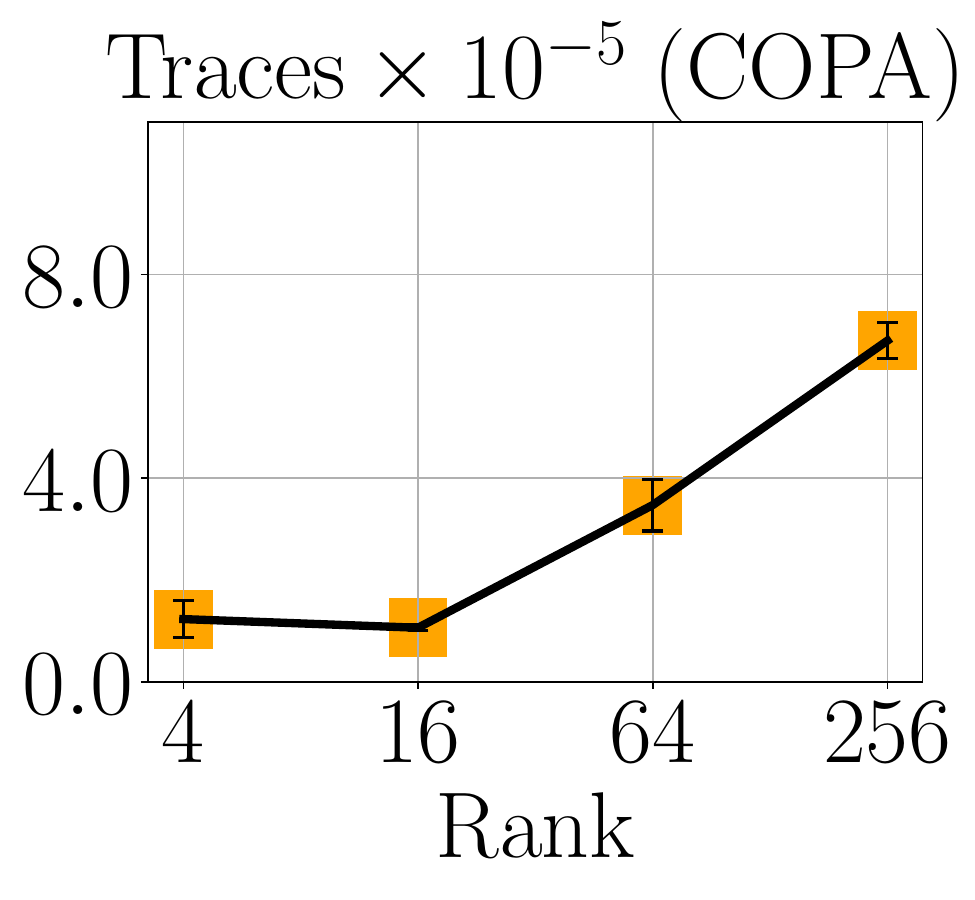}
    \end{subfigure}
    \subcaption{Varying the rank of QLoRA}\label{fig_vary_rank}
    \end{subfigure}
    \begin{subfigure}[b]{0.33\textwidth}
    \centering
    \begin{subfigure}[b]{0.490\textwidth}
    \includegraphics[width=\textwidth]{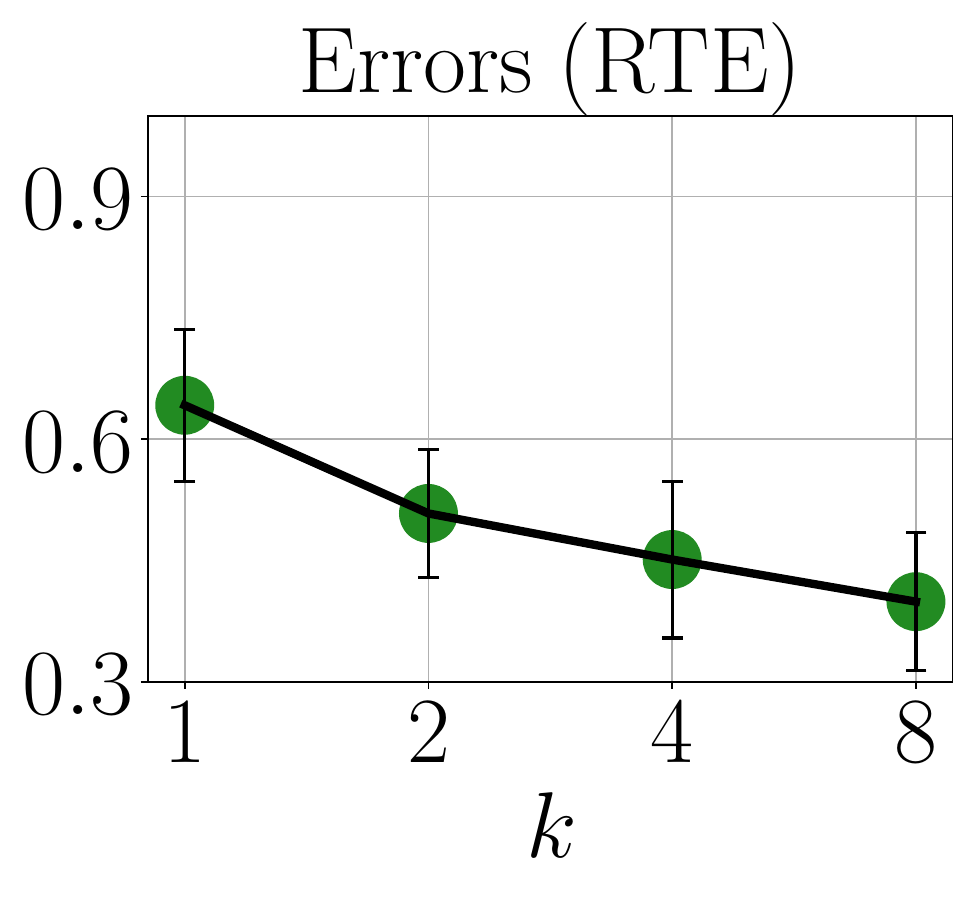}
    \end{subfigure}
    \begin{subfigure}[b]{0.490\textwidth}
    \includegraphics[width=\textwidth]{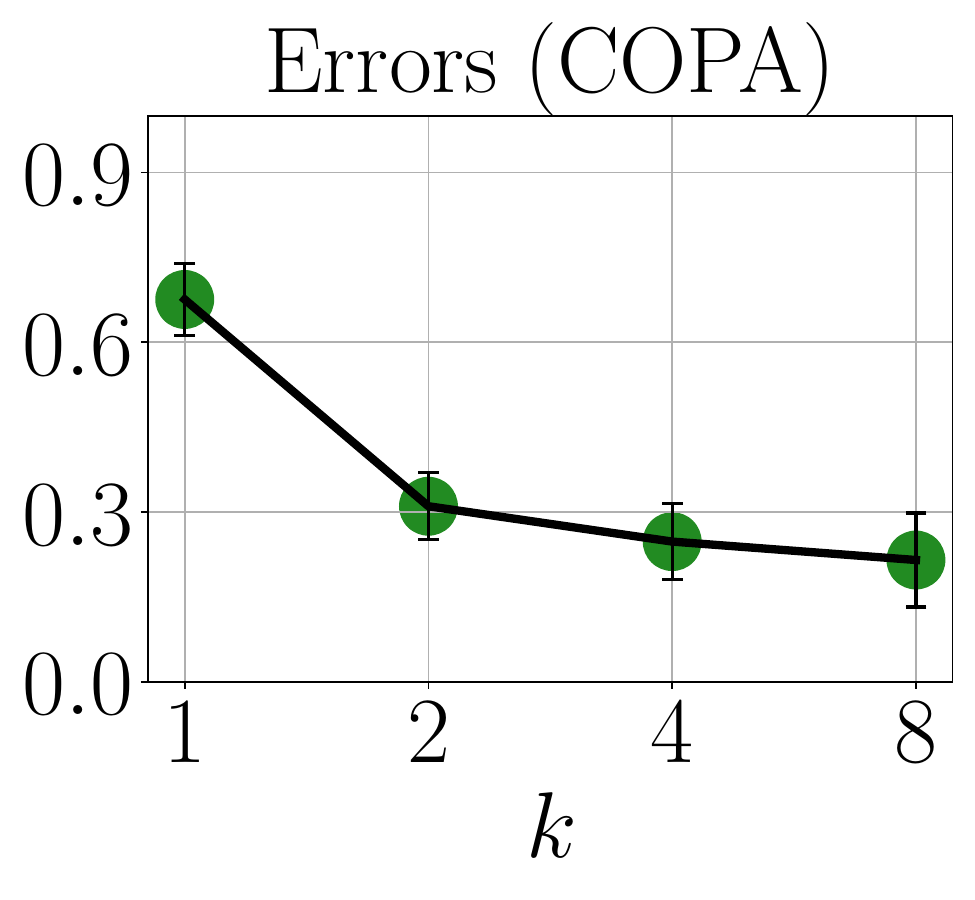}
    \end{subfigure}\vfill
    \begin{subfigure}[b]{0.490\textwidth}
    \includegraphics[width=\textwidth]{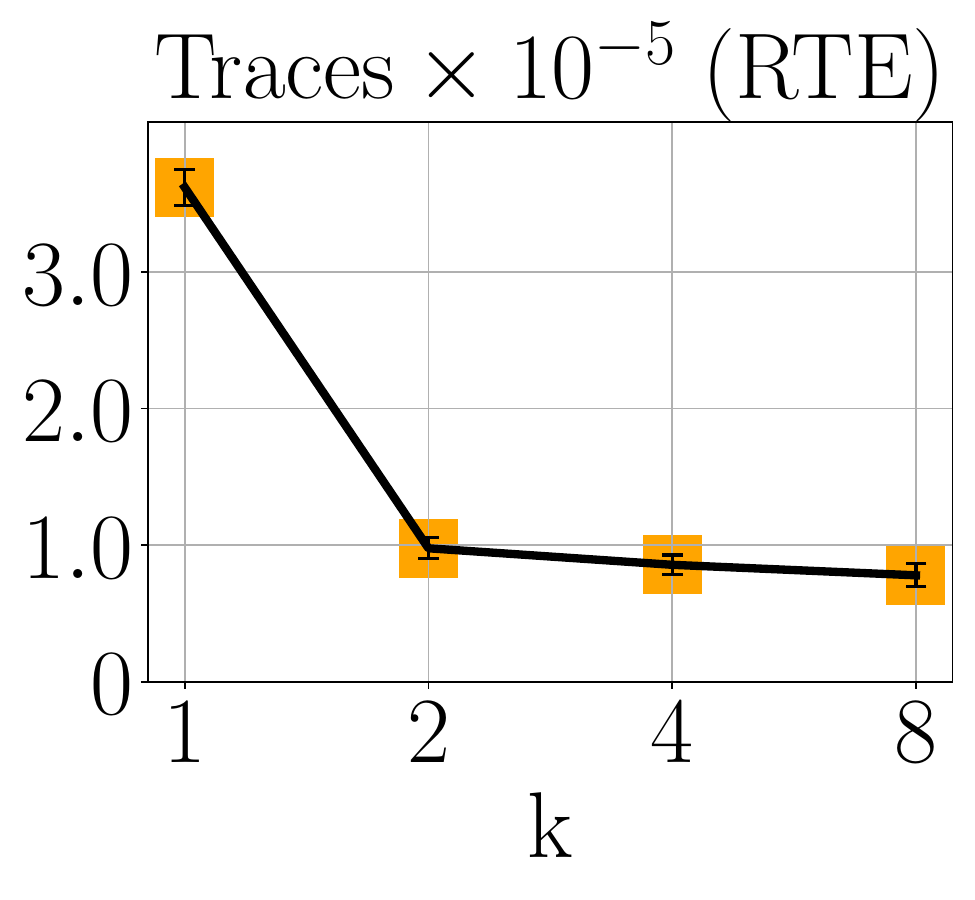}
    \end{subfigure}
    \begin{subfigure}[b]{0.490\textwidth}
    \includegraphics[width=\textwidth]{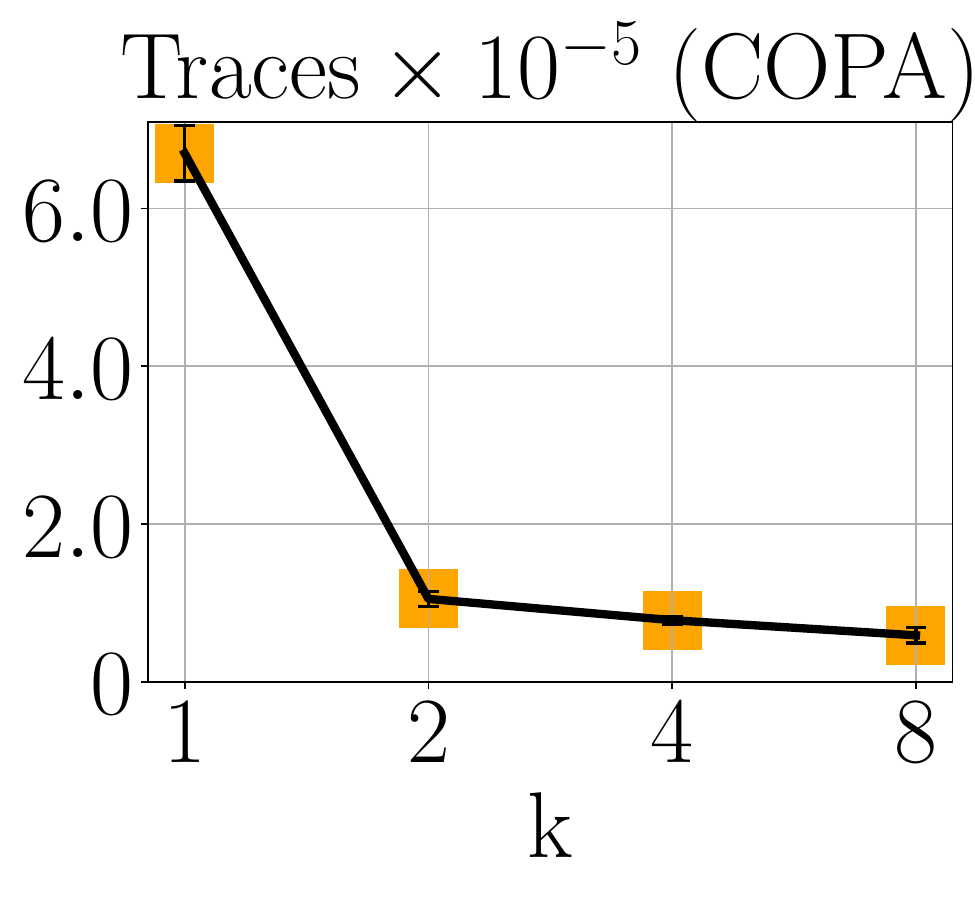}
    \end{subfigure}
    \subcaption{Varying ensemble size}\label{fig_ensemble_size}
    \end{subfigure}
    \begin{subfigure}[b]{0.33\textwidth}
    \centering
    \begin{subfigure}[b]{0.490\textwidth}
    \includegraphics[width=\textwidth]{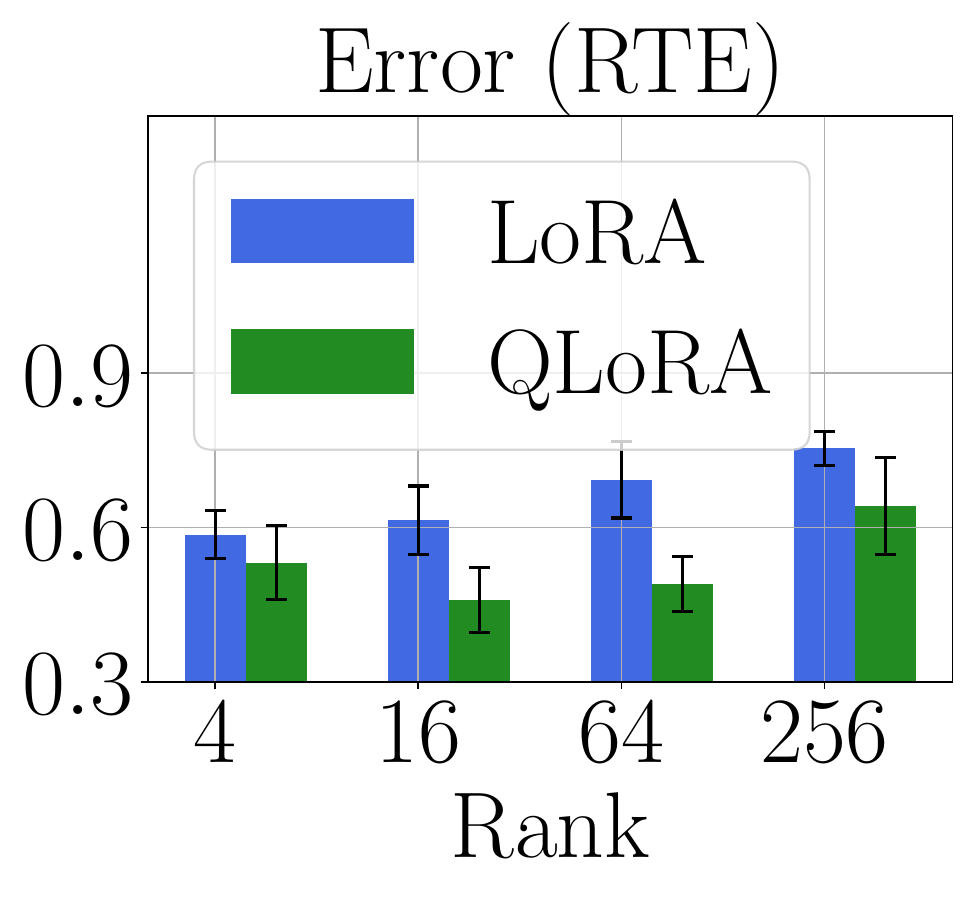}
    \end{subfigure}
    \begin{subfigure}[b]{0.490\textwidth}
    \includegraphics[width=\textwidth]{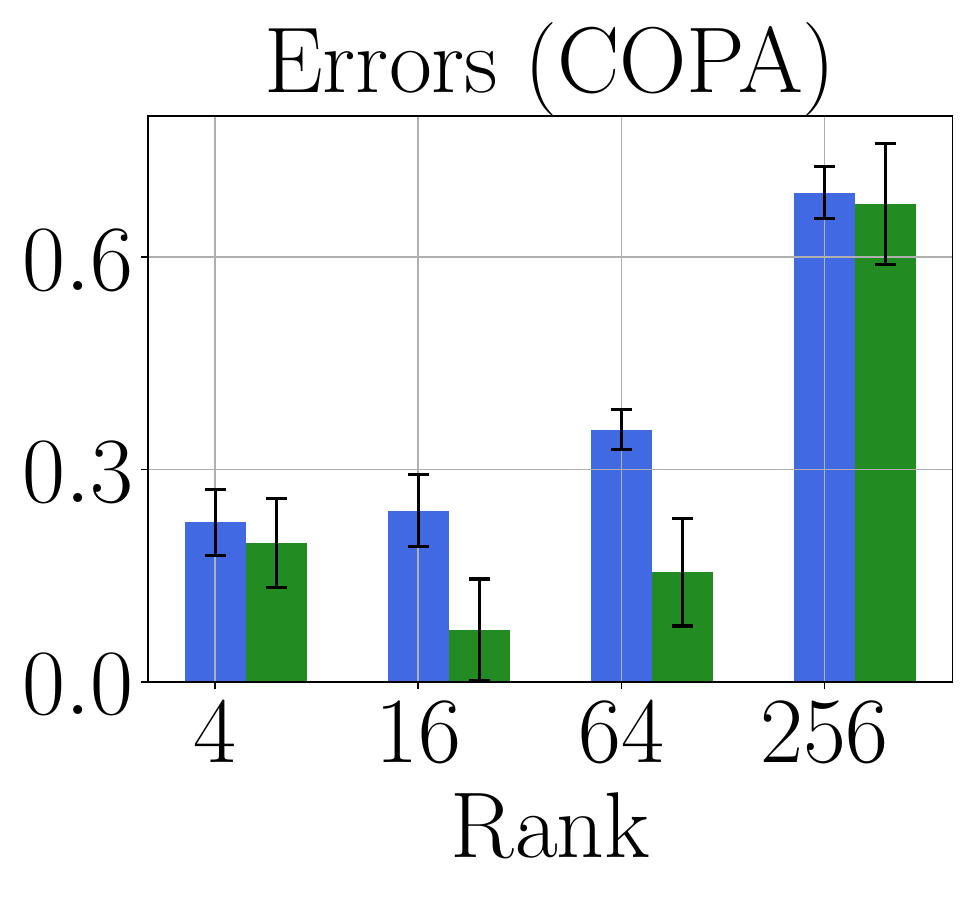}
    \end{subfigure}\vfill
    \begin{subfigure}[b]{0.490\textwidth}
    \includegraphics[width=\textwidth]{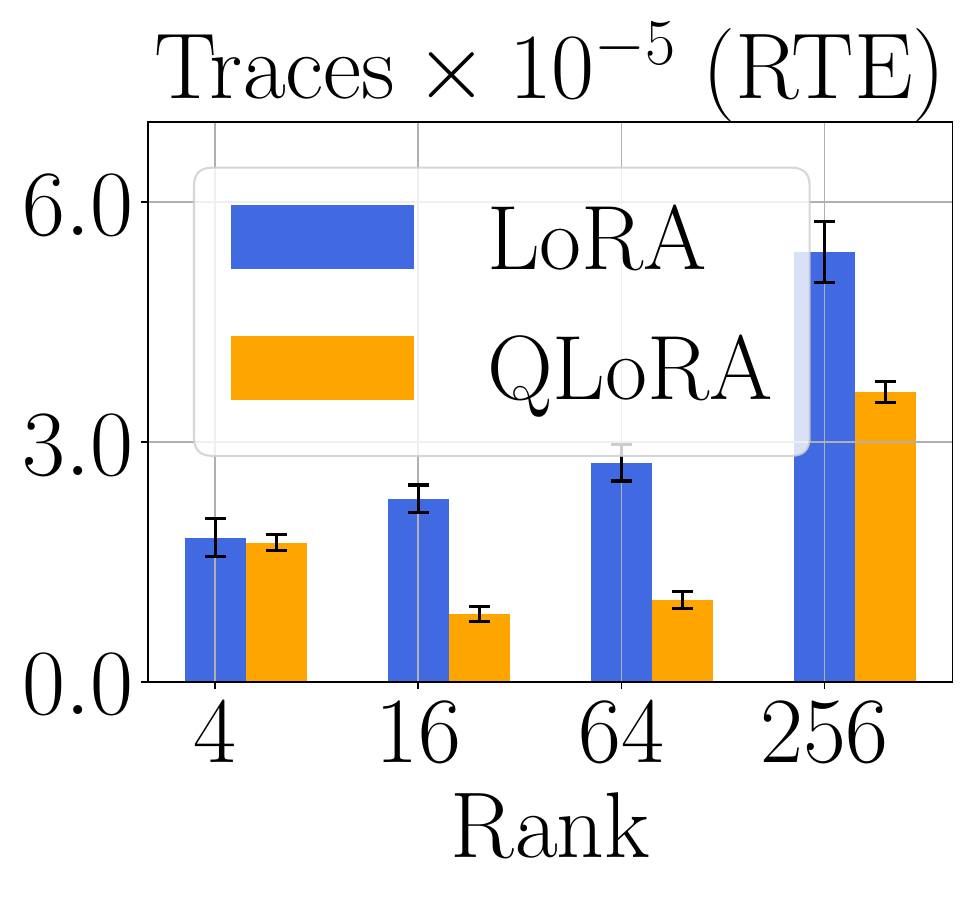}
    \end{subfigure}
    \begin{subfigure}[b]{0.490\textwidth}
    \includegraphics[width=\textwidth]{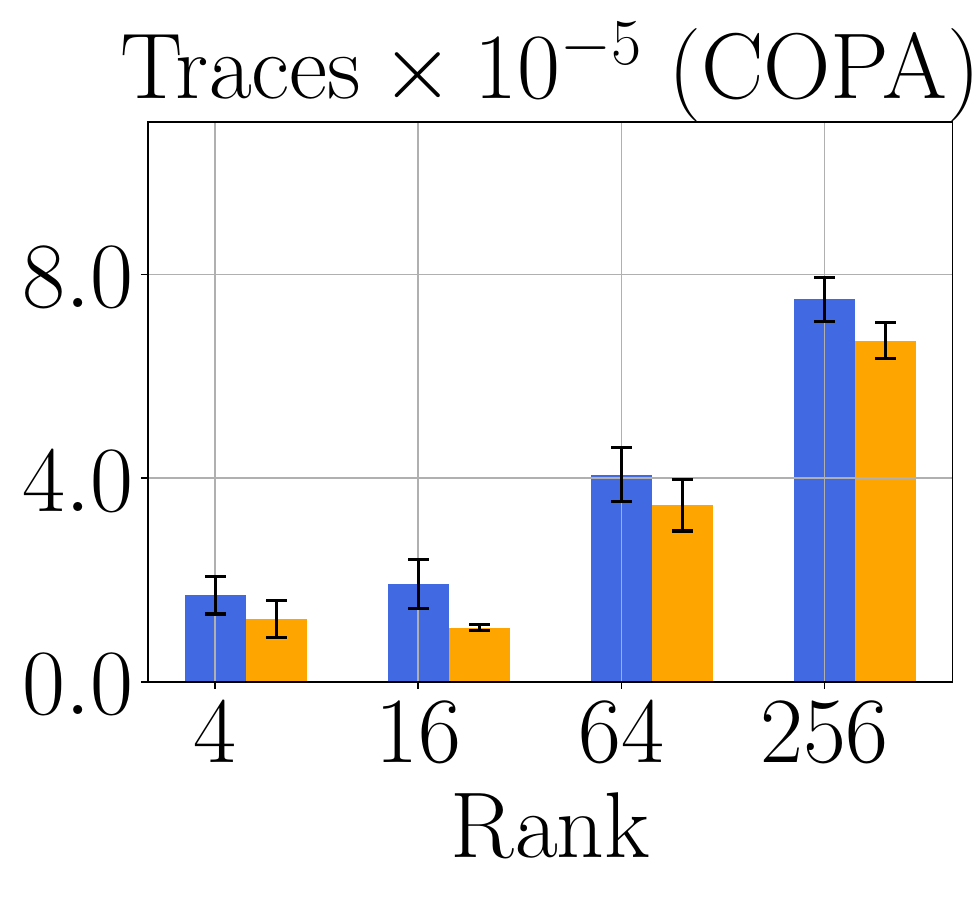}
    \end{subfigure}
    \subcaption{Comparing LoRA and QLoRA}\label{fig_vary_quantize}
    \end{subfigure}%
    \caption{Illustrating the empirical generalization errors and sharpness measures with respect to QLoRA weights.
    \eqref{fig_vary_rank} Smaller adapters with a rank of $16$ achieve the lowest generalization errors. Additionally, the Hessian trace values correlate with generalization errors, suggesting that smaller adapters tend to converge to flatter minima.
    \eqref{fig_ensemble_size} An ensemble of $k$ adapters leads to lower generalization errors and Hessian traces. Here, we fix the sum of the dimensions of $k$ adapters to be equal to $256$.
    \eqref{fig_vary_quantize} QLoRA, which is trained on a quantized base model, yields lower generalization errors and Hessian trace values compared to LoRA.}\label{fig_generalization_errors}
\end{figure*}

Next, we analyze the generalization behavior of low-rank adaptations and their ensembles. Our findings reveal that relatively small adapters often achieve the best generalization performance. Additionally, ensembles of adapters yield lower generalization errors than individual adapters. We also observe that adapters trained on quantized models exhibit smaller generalization errors compared to those trained on full-precision models.

In more detail, we evaluate a sharpness measure based on the trace of the Hessian of the loss.
Previous work has shown that this Hessian-based measure closely correlates with empirical generalization errors \cite{ju2022robust}, which is derived from the PAC-Bayes generalization framework \cite{neyshabur2017pac}.
Specifically, we evaluate the trace of the loss's Hessian with regard to the adapter weights. Then, we compute the maximum of the trace over the training data points. 
A smaller trace value suggests a flatter loss landscape.
We measure the empirical generalization error, defined as the difference between the test loss and the training loss of a fine-tuned model.
{Results for evaluating the largest eigenvalue of the loss Hessian as the sharpness measure are quantitatively similar.}

For a base fine-tuning method, we vary the rank of adapter weights between 4, 16, 64, and 256 in fine-tuning Llama-3-8B. For adapter ensembles, we set a total rank of $k$ adapters as 256. Then, we fine-tune $k$ individual adapters with $256 / k$ rank and ensemble them through weighted averaging. We vary $k$ between 2, 4, and 8 (beyond which the ensemble exceeds the maximum memory limit). Additionally, we compare the generalization errors of fine-tuning adapters on the full-precision model (LoRA) and the 4-bit quantized model (QLoRA). We fine-tune LoRA adapters on two NLP tasks, including RTE and COPA. 

\hypertarget{para_varying_rank}{$\diamondsuit$}
Figure \ref{fig_vary_rank} shows the results of varying the rank of LoRA and QLoRA. We find that using a small-sized adapter with a rank of $16$ typically achieves the smallest generalization error. Moreover, the Hessian-based sharpness measure correlates with the empirical generalization errors, suggesting that a smaller adapter tends to find a fatter minimum.
We find that the results are consistent for other datasets and for adapter tuning, which are shown in Appendix \ref{sec_additional_generalization_measures}.

\hypertarget{para_varying_ensemble_size}{$\diamondsuit$}
Next, shown in Figure \ref{fig_ensemble_size}, for ensembles of $k$ adapters, we find that the ensemble of two adapters reduces the generalization errors by 44\% and the sharpness measure by 74\% on average. Increasing $k$ further reduces the generalization errors. Similarly, the Hessian-based sharpness measure also correlates with generalization errors. This justifies the improved generalization performance of ensembles of adapters in our previous experiments. 

\hypertarget{para_comparing_qlora}{$\diamondsuit$}
Lastly, shown in Figure \ref{fig_vary_quantize}, we also notice that QLoRA, which is trained on the quantized model, yields lower generalization errors than those on the full precision language model. The generalization error and Hessian traces of QLoRA are $34\%$ and $51\%$ lower than LoRA on average. 

\section{Related Work}\label{sec_related}

Parameter-efficient fine-tuning adapts LLMs by either fine-tuning a small subset of parameters or introducing new trainable parameters. %
Adapter tuning \cite{houlsby2019parameter} inserts two feedforward layers per transformer layer, with one for down-projection and another for up-projection of the hidden representations. Prefix tuning \cite{li2021prefix,lester2021power} involves prepending small, continuous task-specific soft prompts to input embeddings. LoRA \cite{hu2021lora} optimizes a low-rank decomposition of parameter updates with regard to pretrained model weights. Building on the individual tuning methods,  \citet{chen2023parameter} propose an automatic parameter allocation algorithm to assign individual tuning methods to different parameter groups. Additionally, quantization can be used to reduce the memory costs of LoRA~\cite{dettmers2024qlora} by training on top of a $4$-bit quantized LLM.
LQ-LoRA~\cite{guo2023lq} employs mixed-precision quantization to further reduce the memory cost, using integer linear programming to dynamically configure bit allocation. We refer readers to recent surveys for further references \cite{ding2023parameter}. %

Several studies have explored training multiple adapters on top of LLMs. 
AdaMix \cite{wang2022adamix} uses a mixture of adaptation modules per Transformer layer on a single dataset. AdaMix first trains the mixture using routing techniques from mixture-of-expert architectures \cite{fedus2022switch} and combines adapter weights through weight averaging \cite{wortsman2022model}. 
Similarly, there has been work on training a sparse mixture of soft prompts, using a gating mechanism to activate specific prompts per sample.
Mini-ensemble of LoRA adapters \cite{ren2024mini} uses fewer trainable parameters while maintaining a higher rank of LoRA adapters to improve the performance of LoRA. 

In the presence of multiple datasets, there has been work on fitting transformers that learns hyper-networks that generate adapters for every layer, which condition on task, adapter position, and layer within a transformer model. AdapterFusion \cite{pfeiffer2020adapterfusion} uses a group of adapters for multiple prediction tasks, by first learning task-specific adapters and then combining adapters with a cross-attention module to fuse the representations computed from each adapter. %

Different notions of generalization in LLMs have been explored, including out-of-distribution generalization \cite{koh2021wilds}. There has been work on designing meta-learning algorithms for prompt tuning, enabling a single prompt to generalize across multiple datasets. Another relevant notion is cross-lingual generalization. \citet{muennighoff2022crosslingual} study prompted fine-tuning with multilingual instructions for zero-shot learning. \citet{liu2023improving} propose scheduled unfreezing to mitigate catastrophic forgetting in multilingual adapter fine-tuning. %

Supervision and representation strategies for language understanding have been well explored. \citet{kozareva2011class} enhance class-instance labeling using instance-instance graph propagation, showing that instance-level connections improve label coverage. \citet{arora2016latent} explain the linear algebraic structures observed in word embeddings using a latent variable model that links partial mutual information with random walk dynamics. \citet{karamanolakis2024interactive} introduce a framework to combine rule extraction with expert feedback for efficient weak supervision. \citet{min2023recent} survey pretrained language models, outlining fine-tuning, prompting, and generation paradigms along with their strengths and challenges.

A series of works have used gradient similarity to measure task similarity. \citet{xia2024less} propose selecting a subset of instruction data for targeted fine-tuning by computing gradient-based influence scores and selecting data with high cosine similarity to the target task.
\citet{park2023trak} introduce a scalable data attribution method that approximates model prediction losses on other data samples when adding or removing individual training samples. \citet{li2024scalable,li2024scalablek} use gradients from a meta initialization to approximate fine-tuning loss and apply the estimates to select auxiliary tasks to jointly fine-tune with the target task.
Inspired by these works, we conduct the first comprehensive study of linearization in parameter-efficient fine-tuning methods. We also propose an adapter ensemble that combines task clustering with gradient boosting.

\section{Conclusion}\label{sec_conclude}

This paper presents an ensemble method of low-rank adapters for adapting language models across multiple datasets. First, we develop an efficient task affinity grouping algorithm, with a first-order approximation for estimating task affinities and a clustering step to partition tasks into groups. Then, we construct an ensemble for groups of tasks, consisting of adapters fine-tuned on each group with additional boosting steps. Our method consistently improves fine-tuning performance with minimal computational overhead. Lastly, we analyze the sharpness measures of low-rank adapters.

\section*{Acknowledgments}

Thanks to the anonymous referees and the action editor for their constructive feedback.
The work of D. Li, Z. Zhang, and H. Zhang is partly supported by NSF award IIS-2412008.
D. Li was also partially funded by a PhD fellowship from JPMorgan Chase \& Co. Any views or opinions expressed herein are solely those of the authors listed, and may differ from those expressed by JPMorgan Chase \& Co. or its Affiliates.

\section*{Limitations}
Our approach requires full access to model weights and gradients. Estimating fine-tuning performance without direct access to internal parameters remains an open question.  This could involve exploring gradient-free optimization methods for closed-source models like GPT-4 and Gemini. Moreover, while our study focuses on multiple fixed datasets, other scenarios require dynamically adapting to new incoming tasks, such as in continual learning. Extending our work to actively manage a dynamic set of adapters could be a promising direction. Besides, our analysis of sharpness measures for generalization performance leaves open the question of their applicability to out-of-distribution generalization or adversarial robustness in NLP models, which may be worth considering in future work.

\section*{Broader Implications}
This paper examines the problem of adapting language models to multiple datasets. While the use of language models may have potential societal consequences in the future, there are no specific concerns arising from our work. Due to the technical nature of this paper, there are no direct implications for potential negative impacts.

\bibliography{ref}
\appendix
\clearpage

\section{Omitted Details about Our Approach}\label{app_approach}

\textbf{Notations.} %
We provide a list of mathematical notations for reference:
\begin{itemize}
    \item $S$: A subset of $\set{1, 2, \dots, n}$.
    \item $f_i(S)$: {The performance of a model fine-tuned on tasks in $S$, evaluated on task $i$}.
    \item $\theta^\star$: {The vectorized model weights of the pretrained initialization}.
    \item $\hat\theta_S$: {The vectorized model weights fine-tuned on a subset of tasks $S$}.
    \item $h_{\theta^\star}(s, y)$: Model output given an input pair $s, y$.
    \item $\nabla h_{X}(s, y)$: {Vectorized gradients of model output with respect to model weights $X$}.
    \item $T_{i,j}$: {The average performance of $f_i(S)$ over multiple subsets $S$ that include task $i$}, for every $i = 1, 2, \dots, n$.
    \item $j_i$: The index of the group chosen at step $i$ of the boosting procedure. 
    \item $G_{j_i}$: The group of tasks chosen at step $i$ of the boosting procedure. 
    \item $t_{j_i}$: The number of adapters with in the group $G_{j_i}$ fit in the gradient boosting procedure. 
\end{itemize}

\subsection{Clustering algorithms}

We now describe a clustering algorithm to partition the $n$ tasks into $k$ disjoint subsets.
Given an $n$ by $n$ task relevance score matrix $T$, we will find a clustering that maximizes the average density of all clusters.
Concretely, let $C_1, \dots, C_k$ be a disjoint partition of $[n]$.
Let $v_1, \dots, v_k$ be a $0$-$1$ vector indicating whether a task is in one cluster.
The average density of this clustering can be written as:
\begin{align*}
   \frac 1 k \sum_{i=1}^{k} \frac{v_i^{\top} T v_i}{v_i^{\top} v_i}. 
\end{align*}
This integral objective is NP-hard to optimize in general.

We design a Semi-Definite Programming (SDP) relaxation and then round the SDP solution to a clustering.
Let us denote the assignment variables as an $n \times k$ matrix $V$, such that each entry $V_{i,j}$ indicates whether a task $i$ belongs to a cluster $j$, for every $i = 1, \dots, n$, $j = 1, \dots, k$.
Moreover, let the $j$th column of $V$, which is the characteristic vector of the $j$-th cluster, be denoted as $v_j$.
Under this assignment, the sum of $V_{i, j}$ across any task $i$ must be one, as we allow one task to be assigned in a single group. By contrast, the sum of $V_{i, j}$ across $C_j$ is the number of tasks assigned to  $C_j$, which is at least one.

Let $e$ denote the all-ones vector. We state an integer program to maximize the average density of all $k$ clusters as follows
\begin{align*}
    \max_{V\in\real^{n\times k}}~~&  \BigInner{T}{ \frac 1 k\sum_{j=1}^k \frac{v_j v_j^{\top}}{v_j^{\top}v_j}} \nonumber \\
    & Ve = e, \sum_{i=1}^n V_{i,j} \geq 1, \text{ for } 1 \leq j \leq k \nonumber \\
    & V_{i, j} \in \set{0, 1}, \text { for } 1 \le i\le n, 1 \le j\le k.
\end{align*}
Note that $v_iv_i^{\top}$ is a rank-one semidefinite matrix.
Let us denote the sum of them (normalized by $v_i^{\top}v_i$) as the following new variable
\begin{align*}
    X = \sum_{j=1}^k \frac{v_j v_j^{\top}}{v_j^{\top}v_j}.
\end{align*}
$X$ has rank $k$ since it is the sum of $k$ rank-$1$ matrices, and the $v_i$'s are orthogonal to each other. Additionally, its trace is equal to $k$ because the trace of $\frac{v_j v_j^{\top}}{v_j^{\top}v_j}$ is one for any $j$.
Second, one can verify that the entries of every row of ${X}$ sum up to one.
Removing the $0$-$1$ integer constraint and regularizing the number of clusters, we derive the problem as
\begin{align*}
    \max_{X\in\real^{n\times n}}~~& \inner{T}{X} + \lambda \tr[X] \\
        & X e = e \\ %
        & X \geq 0,  X \succeq 0.
\end{align*}
where $\lambda$ is a hyperparameter for regularization. This leads to a convex program, which can be solved efficiently.

Given a solution of the SDP, denoted as $\hat X$, the last step is to round $\hat X$ into an integer solution.
We set a threshold $\lambda$ such that if $\hat X_{u,v} \ge \lambda$, tasks $u$ and $v$ are assigned to the same cluster.
In the experiments, we set $\lambda$ as $c/n$ for a constant $c\ge 1$, since $\hat X_{u,v}$ should be $\frac{1}{|C_i|}$ when they are in the same cluster with $|C_i|< n$.
Thus, the intra-cluster distance must always be at least $\lambda$ with the assignment.

\subsection{Boosting algorithms}

In each step of our boosting procedure, it picks the group with the largest training error, denoted as $G_{j_i}$.  Then, compute the negative gradient w.r.t. current model predictions for each sample in the group $(s, y) \in \cD_{G_{j_i}}$:
\[ -  \frac{\partial \cL}{\partial h(s, y)} = 1 - p, \]
where $p$ is the current prediction probability for the correct class label of sample $i$. 
Then, we fit an adapter $\theta_{j_i}^{(t+1)}$ to the combined data set of $\cD_{G_{j_i}}$, mapping the input sample $s$ to the negative gradients $1-p$ in regression. 
Update the current ensemble 
$h_{j_i}^{(t+1)}(s, y) = h_{j_i}^{(t)}(s, y) + \eta h_{\theta_{j_i}^{(t+1)}}(s, y)$.

After multiple steps, the algorithm returns the final ensemble for one group: 
\[ h(s, y) = h_{\theta_{j_i}^{(0)}}(s, y) + \eta \sum_{t=1}^{b} h_{\theta_{j_i}^{(t)}}(s, y). \]
We use $\eta$ as $0.1$. 
Note that we use one group as an example. We apply the boosting procedure to $m$ groups in our approach. 

\paragraph{Extension to AdaBoost:} In a similar sense, we can apply AdaBoosting to boost adapters. 
At each iteration, the sample weights are modified, and a new adapter is trained on the reweighted training samples. 
Denote the error rate of $h^{(t)}$ at iteration $t$ as $err^{(m)}$. 
Denote $\alpha^{(t)} = \log \frac{1 - err^{(t)}}{err^{(t)}}$. 
The training samples that are misclassified by the classifier $h^{(t-1)}$ have their weights increased exponentially by $err^{(t)}$:  
\[ w_i^{(t)} \leftarrow w_i^{(t-1)} \exp\Big(\alpha^{(t)} \mathbf{1}\big(y_i \neq h_{\theta^{(m)}}(s_i, y_i)\big)\Big) \]
In our setup, weights can be assigned to separate tasks instead of single tasks. 
We use parameter-efficient fine-tuning to fit each model, such as LoRA and Adapters. 
At each iteration, a new model is fit on a reweighted version of the training dataset. Similarly, the first-order approximation can be applied to this case by solving a reweighted version of logistic regression on gradients. 
Therefore, we can also leverage Algorithm \ref{alg_gradient_estimation} to fit the adapter weights in each new model quickly. 

\subsection{Computation and memory costs}

The computation cost of our approach involves: %
\begin{itemize}
    \item $O(n)$ gradient evaluation of all the tasks and solving logistic regression on $M$ random subsets. 
    \item Fine-tuning $m$ adapters, each on a cluster of tasks. The combination of $m$ task groups is $n$ tasks.
    \item Applying $b$ iterations of gradient boosting on the adapters, which computes $O(n)$ gradients of the existing model's outputs.
\end{itemize}
Thus, Algorithm \ref{alg_boosting} involves $O(n)$ computation in the number of forward and backward passes. This is comparable to training a single multitask model on all tasks. We report the exact computation cost of our approach in experiments. 

As for the memory, Algorithm \ref{alg_boosting} uses the memory for maintaining $m + b$ adapter models and one base model.

\section{Omitted Experiments}\label{app_exp}

\subsection{List of models and datasets}\label{app_dm}

We experiment with the following models: \href{https://huggingface.co/meta-llama/Llama-3.2-1B}{Llama-3.2-1B}, \href{https://huggingface.co/meta-llama/Llama-3.2-3B}{Llama-3.2-3B}, \href{https://huggingface.co/EleutherAI/gpt-j-6b}{GPT-J-6B}, \href{https://huggingface.co/meta-llama/Llama-3.1-8B}{Llama-3.1-8B}, \href{https://huggingface.co/meta-llama/Llama-2-13b-hf}{Llama-2-13B}, and \href{https://huggingface.co/meta-llama/CodeLlama-34b-Instruct-hf}{CodeLlama-34B-Instruct}.

The datasets used in our experiments include \href{https://super.gluebenchmark.com/}{SuperGLUE}, \href{https://rowanzellers.com/hellaswag/}{HellaSwag}, \href{https://cs.rochester.edu/nlp/rocstories/}{Story Cloze}, \href{https://winogrande.allenai.org/}{WinoGrande} and the \href{https://huggingface.co/datasets/flwrlabs/shakespeare}{Shakespeare} dataset.
We refer readers to their respective web pages for a detailed description of the dataset statistics. Due to computational constraints, we use 50\% of the training set for training and sample 10\% of it for validation. We report the test accuracy on the development set provided by the datasets. 

\subsection{Results for first-order approximation}

We report the first-order approximation error for the outputs of fine-tuned language models by varying the fine-tuning distance from 0.05\% to 0.25\% across six pretrained language models in Table \ref{tab_approximation_pretrained_models}.

We report the correlation score between the estimated and true fine-tuning performances. We observe that our approach yields a correlation score up to $0.6$, evaluated on the results of $50$ randomly sampled subsets of size $3$, across three Llama models of size up to $8$ billion.
The results are evaluated across $d$ between $200, 400,$ and $800$.

\begin{table*}[t!]
\centering
\caption{We evaluate the first-order approximation errors within the fine-tuning region close to the model pre-trained initialization. 
We measure the fine-tuned distance as $\frac{\normFro{X-\theta^\star}}{\normFro{\theta^\star_{\text{Full}}}}$. We evaluate the RSS
error between $h_X(s, y)$  and the first-order approximation using the pretrained model weights $\theta^\star$.
To ensure the statistical significance of these results, we report the average over $50$ random task subsets.
}\label{tab_approximation_pretrained_models}
{\small\begin{tabular}{@{}ccccccccc@{}}
\toprule
\textbf{Distance}/RSS & {\textbf{Llama-3.2-1B}} & {\textbf{Llama-3.2-3B}} & {\textbf{GPT-J-6B}} & {\textbf{Llama-3.1-8B}} & {\textbf{Llama-2-13B}}   \\\midrule
& \multicolumn{6}{c}{Full-precision pretrained model with LoRA fine-tuning}  \\ \midrule
0.05\% &    $9.1_{\pm 3.1}\times10^{-4}$ & $6.7_{\pm 3.3}\times10^{-4}$ & $4.5_{\pm 0.8}\times10^{-5}$ & $3_{\pm 0.3}\times10^{-5}$ & $2.5_{\pm 1.1}\times10^{-3}$  \\
0.10\% &    $1.9_{\pm 0.4}\times10^{-3}$ & $9.1_{\pm 3.9}\times10^{-4}$ & $4.9_{\pm 0.9}\times10^{-5}$ & $6_{\pm 0.9}\times10^{-5}$ & $8.2_{\pm 0.4}\times10^{-3}$  \\
0.15\% &    $3.4_{\pm 1.2}\times10^{-3}$ & $1.6_{\pm 0.9}\times10^{-3}$ & $5.2_{\pm 1.3}\times10^{-5}$ & $3_{\pm 0.6}\times10^{-4}$ & $3.1_{\pm 0.0}\times10^{-3}$  \\
0.20\% &    $5.0_{\pm 1.1}\times10^{-3}$ & $2.3_{\pm 0.8}\times10^{-3}$ & $1.3_{\pm 1.0}\times10^{-4}$ & $4_{\pm 0.4}\times10^{-4}$ & $3.3_{\pm 2.2}\times10^{-3}$  \\
0.25\% &   $5.4_{\pm 1.0}\times10^{-3}$ & $5.2_{\pm 2.0}\times10^{-3}$ & $1.4_{\pm 0.6}\times10^{-4}$ & $5_{\pm 0.4}\times10^{-4}$ & $2.3_{\pm 1.2}\times10^{-2}$  \\ \midrule
& \multicolumn{6}{c}{4-bit quantized pretrained model with Quantized LoRA (QLoRA) fine-tuning} \\ \midrule
0.05\% &    $1.7_{\pm 0.5}\times10^{-4}$ & $2.0_{\pm 0.9}\times10^{-4}$ & $1.0_{\pm 0.4}\times10^{-4}$ & $1.2_{\pm 1.1}\times10^{-3}$ & $2.9_{\pm 0.3}\times10^{-3}$  \\
0.10\% &    $5.6_{\pm 2.8}\times10^{-4}$ & $2.1_{\pm 1.0}\times10^{-4}$ & $1.9_{\pm 0.2}\times10^{-4}$ & $1.4_{\pm 0.4}\times10^{-3}$ & $3.3_{\pm 0.7}\times10^{-3}$  \\
0.15\% &    $1.0_{\pm 0.5}\times10^{-3}$ & $6.2_{\pm 1.9}\times10^{-4}$ & $2.5_{\pm 2.3}\times10^{-4}$ & $5.3_{\pm 4.2}\times10^{-3}$ & $4.6_{\pm 0.4}\times10^{-3}$  \\
0.20\% &    $2.0_{\pm 0.5}\times10^{-3}$ & $1.1_{\pm 0.5}\times10^{-3}$ & $2.8_{\pm 0.6}\times10^{-4}$ & $3.7_{\pm 2.4}\times10^{-3}$ & $4.9_{\pm 1.0}\times10^{-3}$  \\
0.25\% &   $2.2_{\pm 0.9}\times10^{-3}$ & $1.7_{\pm 0.7}\times10^{-3}$ & $4.3_{\pm 2.5}\times10^{-4}$ & $1.3_{\pm 0.9}\times10^{-2}$ & $6.8_{\pm 0.7}\times10^{-3}$  \\ 
\midrule
\textbf{Distance}/RSS & \textbf{Llama-3.2-1B} & {\textbf{Llama-3.2-3B}} & {\textbf{GPT-J-6B}} & {\textbf{Llama-3-8B}} & {\textbf{Llama-2-13B}} &   \\\midrule
& \multicolumn{6}{c}{Full-precision pretrained model with adapter tuning}  \\ \midrule
0.05\% & $8.9_{\pm 3.0}\times10^{-3}$ & $ 1.4_{\pm 0.1}\times10^{-2}$ & $ 3.5_{\pm 0.8}\times10^{-3}$ & $ 1.3_{\pm 0.2}\times10^{-6}$ & $ 7.7_{\pm 1.1}\times10^{-7}$  \\
0.10\% & $1.5_{\pm 1.0}\times10^{-2}$ & $ 1.9_{\pm 0.8}\times10^{-2}$ & $ 2.8_{\pm 1.3}\times10^{-3}$ & $ 1.4_{\pm 0.1}\times10^{-6}$ & $ 1.2_{\pm 0.8}\times10^{-6}$  \\
0.15\% & $2.1_{\pm 1.1}\times10^{-2}$ & $ 1.8_{\pm 1.0}\times10^{-2}$ & $ 3.4_{\pm 0.8}\times10^{-3}$ & $ 1.9_{\pm 0.4}\times10^{-6}$ & $ 1.4_{\pm 0.1}\times10^{-6}$  \\
0.20\% & $2.1_{\pm 0.5}\times10^{-2}$ & $ 2.0_{\pm 0.7}\times10^{-2}$ & $ 4.1_{\pm 0.7}\times10^{-3}$ & $ 2.2_{\pm 1.6}\times10^{-6}$ & $ 1.5_{\pm 0.8}\times10^{-6}$  \\
0.25\% & $7.0_{\pm 1.1}\times10^{-2}$ & $ 7.5_{\pm 6.4}\times10^{-2}$ & $ 6.9_{\pm 6.7}\times10^{-3}$ & $ 2.9_{\pm 1.9}\times10^{-6}$ & $ 1.5_{\pm 0.7}\times10^{-6}$  \\ \midrule
& \multicolumn{6}{c}{4-bit quantized pretrained model with quantized adapter (QAdapter) tuning} \\ \midrule
0.05\%   & $7.0_{\pm 1.7}\times10^{-3}$ & $ 1.5_{\pm 0.2}\times10^{-2}$ & $ 6.7_{\pm 0.0}\times10^{-7}$ & $3.0_{\pm 1.6}\times10^{-6}$ & $ 1.8_{\pm 0.3}\times10^{-6}$  \\
0.10\%   & $1.1_{\pm 0.9}\times10^{-2}$ & $ 1.7_{\pm 0.8}\times10^{-2}$ & $ 7.3_{\pm 0.8}\times10^{-7}$ & $3.1_{\pm 0.8}\times10^{-6}$ & $ 2.1_{\pm 1.4}\times10^{-6}$  \\
0.15\%   & $1.1_{\pm 0.9}\times10^{-2}$ & $ 1.8_{\pm 0.6}\times10^{-2}$ & $ 8.0_{\pm 0.5}\times10^{-7}$ & $5.4_{\pm 3.4}\times10^{-6}$ & $ 2.2_{\pm 1.8}\times10^{-6}$  \\
0.20\%   & $1.4_{\pm 0.5}\times10^{-2}$ & $ 1.9_{\pm 0.5}\times10^{-2}$ & $ 8.8_{\pm 1.1}\times10^{-7}$ & $6.1_{\pm 6.4}\times10^{-6}$ & $ 1.8_{\pm 0.4}\times10^{-6}$  \\
0.25\%  & $1.5_{\pm 0.6}\times10^{-2}$ & $ 4.0_{\pm 1.9}\times10^{-2}$ & $ 1.1_{\pm 0.2}\times10^{-6}$ & $9.1_{\pm 9.6}\times10^{-6}$ & $ 2.5_{\pm 0.6}\times10^{-6}$  \\ 
\bottomrule
\end{tabular}}
\end{table*}

\subsection{Results on adapter tuning}

\begin{figure*}[t!]    
    \begin{subfigure}[b]{0.245\textwidth}
    \centering    
    \includegraphics[width=0.995\textwidth]{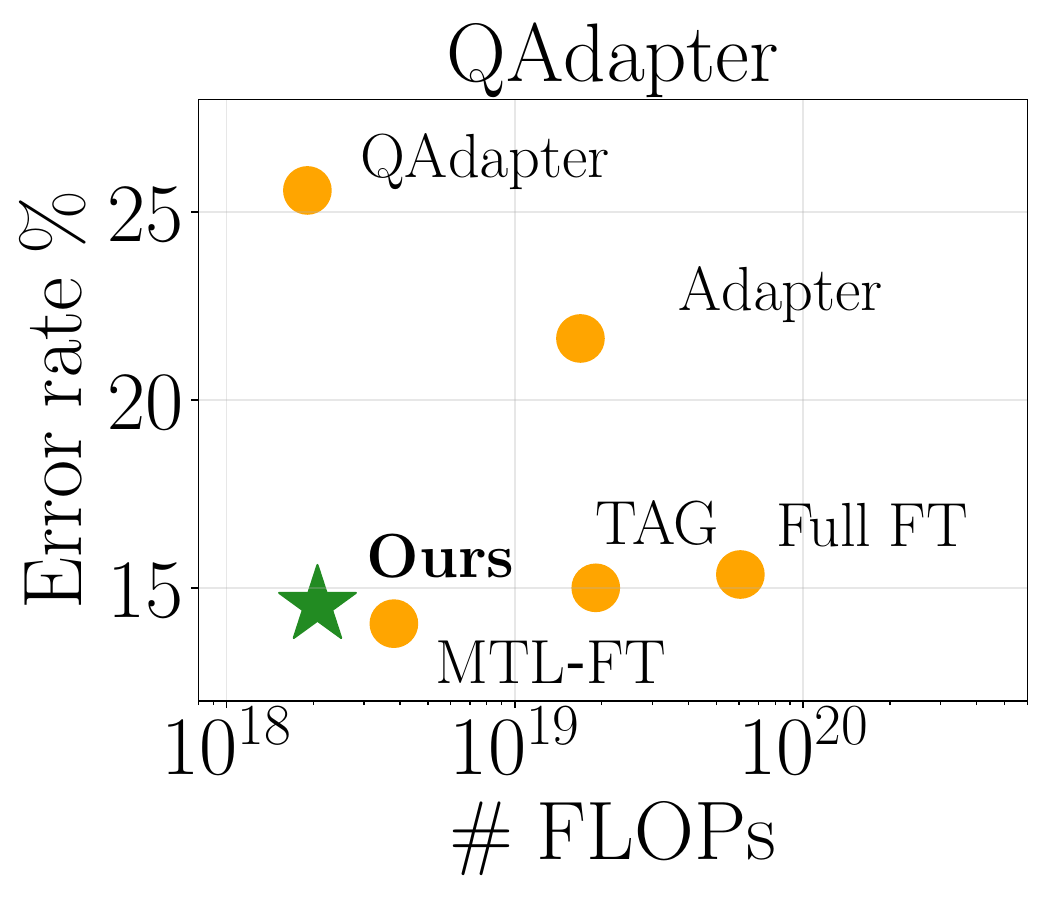}
    \end{subfigure}\hfill
    \begin{subfigure}[b]{0.235\textwidth}
    \centering    
    \includegraphics[width=0.995\textwidth]{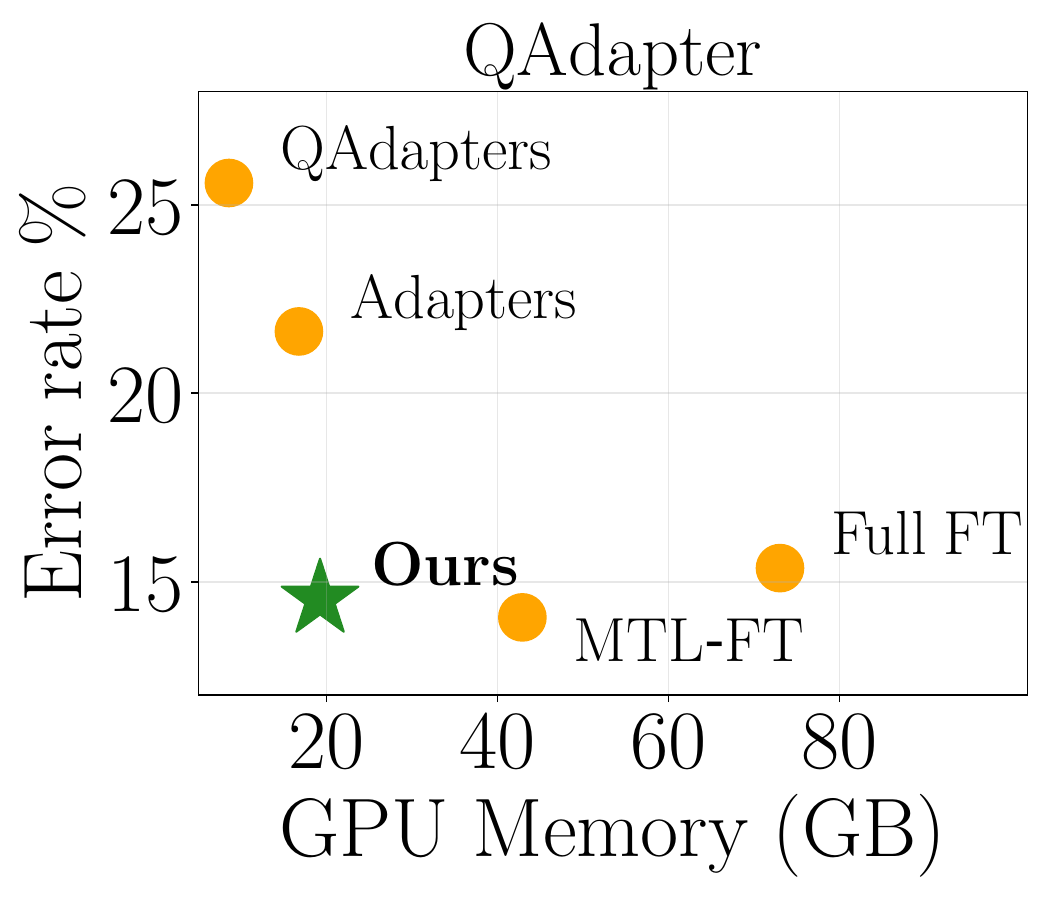}
    \end{subfigure}
    \begin{subfigure}[b]{0.245\textwidth}
    \centering    
    \includegraphics[width=0.995\textwidth]{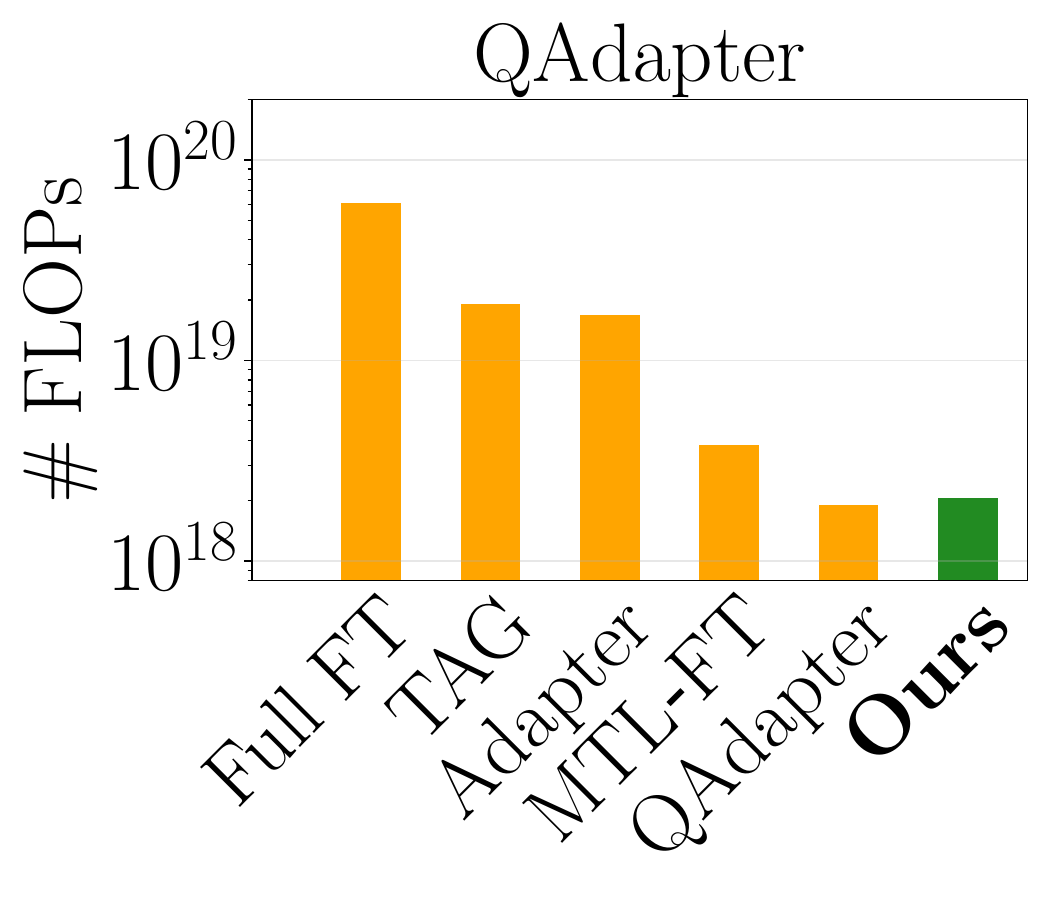}
    \end{subfigure}\hfill
    \begin{subfigure}[b]{0.235\textwidth}
    \centering    
    \includegraphics[width=0.995\textwidth]{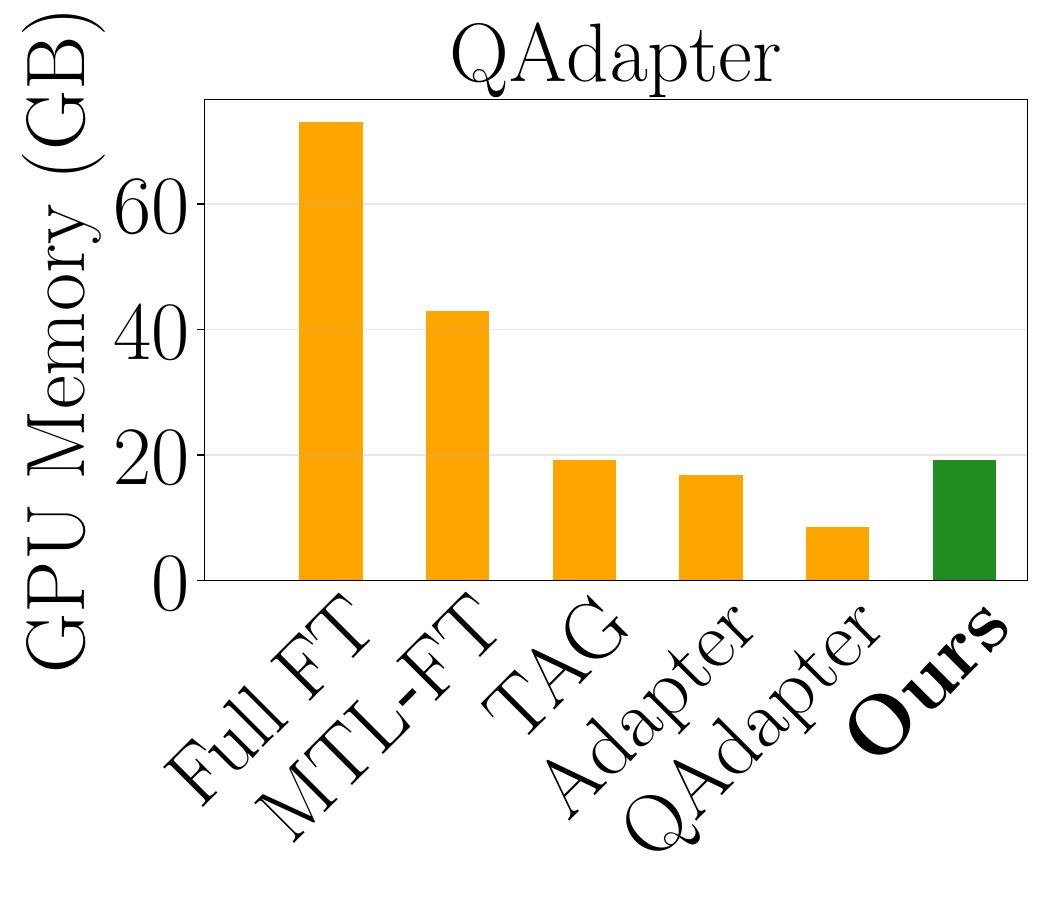}
    \end{subfigure}
    \caption{This figure compares the error rate (one average minus test accuracy), computation cost, and GPU memory across our approach and baselines, for fine-tuning Llama-3-8B on ten NLP tasks with QAdapter. }\label{fig_tradeoff_exp_2}
\end{figure*}

\textbf{Fine-tuning Llama-8B with QAdapter.}
In this section, we report additional evaluation results by using QAdapter as the base fine-tuning protocol in our ensemble method.
Figure \ref{fig_tradeoff_exp_2} presents the results of using QAdapter. Our approach improves QAdapter accuracy by \textbf{9}\%, incurring 8\% more computation and increasing memory from 9GB to 19GB. Compared to MTL-FT and task grouping, it achieves similar accuracy while reducing computation by 46\% and memory from 43GB to 19GB. Against full fine-tuning, it lowers computation by 96\% and memory by {74}\%, while improving test accuracy by \textbf{1}\%.

Next, we report the full comparison in fine-tuning LLama-3-8B on ten NLP tasks using QAdapter and fine-tuning CodeLlama-34B-Instruct using QLoRA in Table \ref{tab_full_results}.

For the hyperparameters, in all our experiments, we fine-tune models with AdamW with a learning rate $2e^{-5}$ for ten epochs. For LoRA, we set the LoRA rank to 4. For Adapters, we set the reduction ratio to 256. These are determined by a hyperparameter search via cross-validation.

\begin{table*}[t!]
\centering
\caption{We report the test accuracy (\%) of our method, as compared with baselines. We also compute the average test accuracy across ten NLP tasks, along with the number of FLOPs and memory usage.  We report the results of fine-tuning CodeLlama-34B-Instruct using QLoRA.  We run each experiment with three random seeds and report the standard deviations.} \label{tab_full_results} %
\resizebox{\textwidth}{!}
{
\begin{tabular}{lcccccccccc|ccc}
\toprule
& BoolQ & CB & COPA & H-SWAG & MultiRC & RTE & Story Cloze & WiC & Winogrande & WSC &
\multirow{5}{*}{\makecell{Average\\Accuracy}} & \multirow{5}{*}{\makecell{Number of\\ FLOPs}} & \multirow{5}{*}{\makecell{GPU \\ Memory}} \\
\# Train & 4,242 & 225 & 360 & 17,957 & 12,259 & 1,120 & 841 & 2,442 & 4,161 & 498\\
\# Valid & 471 & 25 & 40 & 1,995 & 1,362 & 200 & 188 & 270 & 462 & 56\\
\# Test & 3,270 & 56 & 100 & 10,042 & 4,848 & 277 & 1871 & 638 & 1267 & 104\\
\# classes & 2 & 3 & 2 & 4 & 2 & 2 & 2 & 2 & 2 & 2\\ \midrule
QLoRA & 91.9$_{\pm0.7}$ & 96.2$_{\pm0.6}$ & 90.0$_{\pm1.0}$ & 90.7$_{\pm0.9}$ & 89.5$_{\pm0.3}$ & 84.6$_{\pm0.4}$ & 96.9$_{\pm0.6}$ & 74.2$_{\pm0.5}$ & 76.6$_{\pm2.4}$ & 75.5$_{\pm1.2}$ & 86.9$_{\pm 1.6}$ & 1.8$\times 10^{19}$ & 18.6GB \\
MTL-FT & 92.1$_{\pm0.4}$ & 100$_{\pm0.0}$ & 95.0$_{\pm0.5}$ & 96.1$_{\pm0.9}$ & 91.7$_{\pm0.4}$ & 90.7$_{\pm0.4}$ & 98.2$_{\pm0.9}$ & 78.9$_{\pm0.4}$ & 85.1$_{\pm1.1}$ & 82.7$_{\pm1.5}$ & 90.2$_{\pm 0.8}$ & 3.7$\times 10^{19}$ & 65.8GB\\
Ours & 92.1$_{\pm0.4}$ & 100$_{\pm0.0}$ & 94.0$_{\pm1.5}$ & 94.3$_{\pm0.4}$ & 90.7$_{\pm0.4}$ & 88.0$_{\pm0.7}$ & 98.1$_{\pm0.8}$ & 74.1$_{\pm0.6}$ & 84.3$_{\pm1.5}$ & 76.9$_{\pm1.4}$ & 89.9$_{\pm 0.9}$ & 1.9$\times 10^{19}$ & 47.8GB \\
\bottomrule
\end{tabular}
}
\end{table*}

\subsection{Clustering based on gradient similarities}

There has been a line of work on using gradient similarity to measure task similarity. 
In particular, \citet{park2023trak} proposed a scalable data attribution method to quantify the influence of adding or removing one sample on the prediction of the other samples. Their approach also uses JL to reduce the dimension of the gradient features before solving each logistic regression.
\citet{xia2024less} designed a method to select a small subset of instruction data for targeted instruction tuning on a target task. Their method computes gradient-based influence scores by comparing low-dimensional projected gradients from multiple model checkpoints in a warmup training phase. Then, they select data based on cosine similarity to target task gradients.

It should be worth emphasizing that our approach adds very little overhead relative to these gradient similarity-based methods, since we only need to compute the gradients at the base model once. It is an interesting question to further explore the use of gradient similarity as a measure. In our preliminary study, we find that by clustering the tasks with their gradients, the (average) test accuracy remains $0.8\%$ lower than our method.
The results are reported in Table \ref{tab_full_results_gradient_similarity}. 

In terms of how our work complements prior works, first, we have measured the linearization accuracy of parameter-efficient fine-tuning methods, including LoRA, Adapter tuning, quantized LoRA (QLoRA), and QAdapter, on a wide range of LLMs.
Our empirical finding reinforces the belief that, locally, the output of large models can be accurately linearized, as these parameter-efficient fine-tuning methods all stay very close to the base pretrained model. Thus, the approximation error is less than $3\%$ relative to model outputs. This has not been observed in prior works, including the work of \citet{park2023trak}.
Second, we designed an ensemble method to use several low-rank adapters for multiple datasets as opposed to a single (high-rank) adapter for all datasets.
This is complementary to the work of \citet{xia2024less}.
Third, we conducted extensive evaluations of our ensemble method in various multitask settings.
This is again complementary to both of these prior works.

\begin{table*}[t!]
\centering
\caption{We compare our method, which uses gradients to estimate fine-tuning performances, with using gradient similarity for clustering tasks. We report the test accuracies in fine-tuning Llama-3-8B using QLoRA. We run each experiment with three random seeds and report the standard deviations.} \label{tab_full_results_gradient_similarity}
\resizebox{\textwidth}{!}
{
\begin{tabular}{lccccccccccccc}
\toprule
& BoolQ & CB & COPA & H-SWAG & MultiRC & RTE & Story Cloze & WiC & Winogrande & WSC  \\
\midrule
Clustering with gradient similarities & 88.0$_{\pm0.7}$ & 100$_{\pm0.0}$ & 92.0$_{\pm1.0}$ & 92.5$_{\pm0.6}$ & 89.1$_{\pm0.3}$ & 83.0$_{\pm0.2}$ & 96.5$_{\pm0.9}$ & 70.3$_{\pm0.7}$ & 81.5$_{\pm0.9}$ & 79.6$_{\pm0.3}$ \\
Clustering by task affinity scores (Ours) & 89.9$_{\pm0.9}$ & 100$_{\pm0.0}$ & 94.0$_{\pm1.0}$ & 93.5$_{\pm0.3}$ & 89.1$_{\pm0.3}$ & 85.7$_{\pm0.3}$ & 97.1$_{\pm0.1}$ & 69.9$_{\pm0.9}$ & 82.0$_{\pm0.5}$ & 79.8$_{\pm2.2}$\\
\bottomrule
\end{tabular}
}
\end{table*}

\subsection{Sharpness measurements}\label{sec_additional_generalization_measures}

\begin{figure}[h!]
    \begin{subfigure}[b]{0.155\textwidth}
    \centering
    \begin{subfigure}[b]{\textwidth}
    \includegraphics[width=\textwidth]{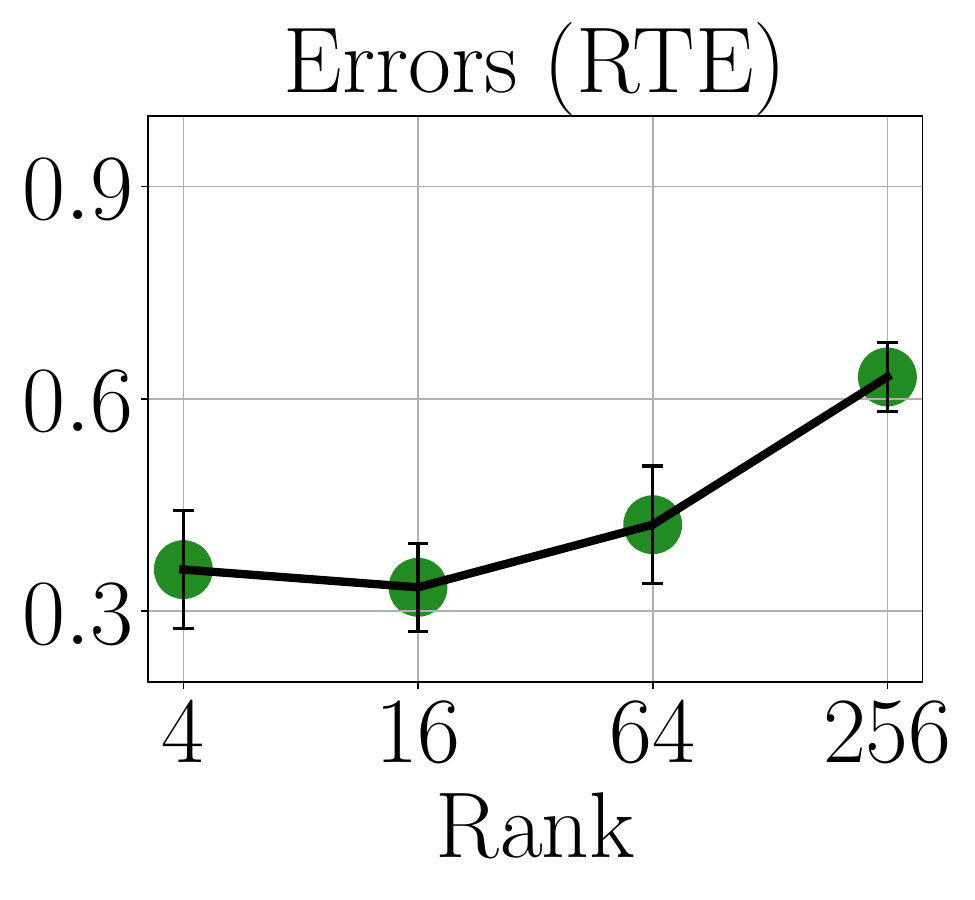}
    \end{subfigure}\vfill
    \begin{subfigure}[b]{\textwidth}
    \includegraphics[width=\textwidth]{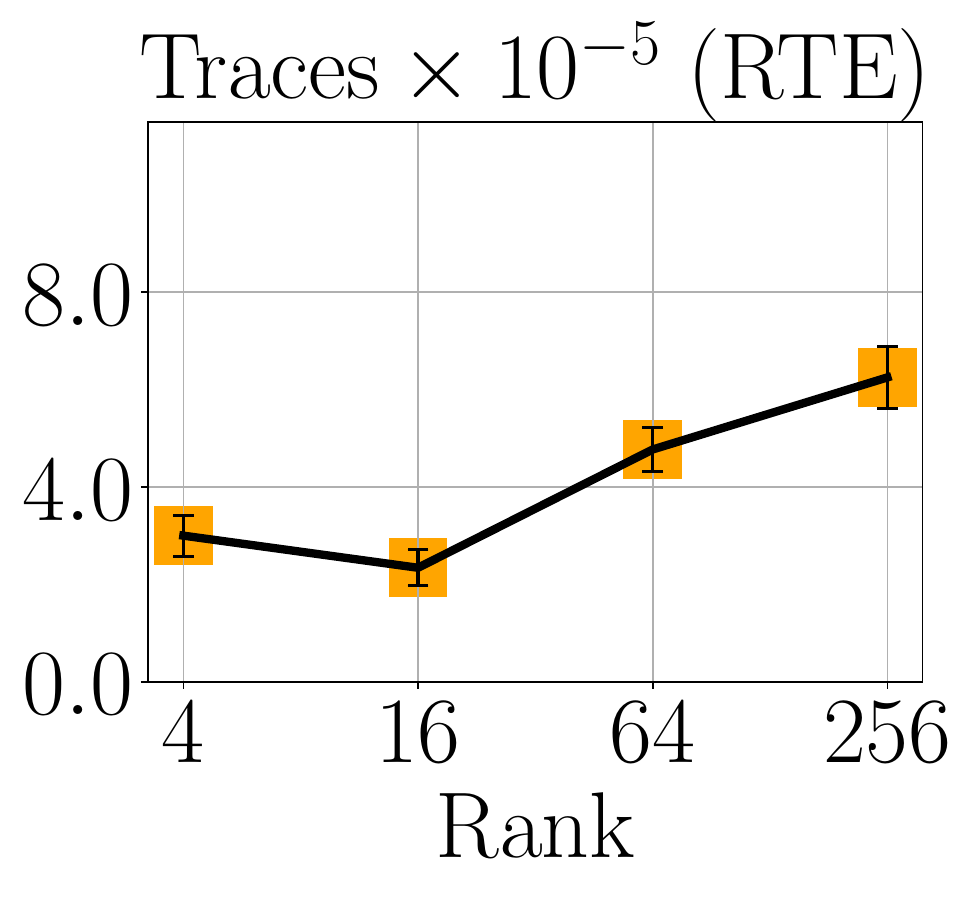}
    \end{subfigure}
    \end{subfigure}
    \begin{subfigure}[b]{0.155\textwidth}
    \centering
    \begin{subfigure}[b]{\textwidth}
    \includegraphics[width=\textwidth]{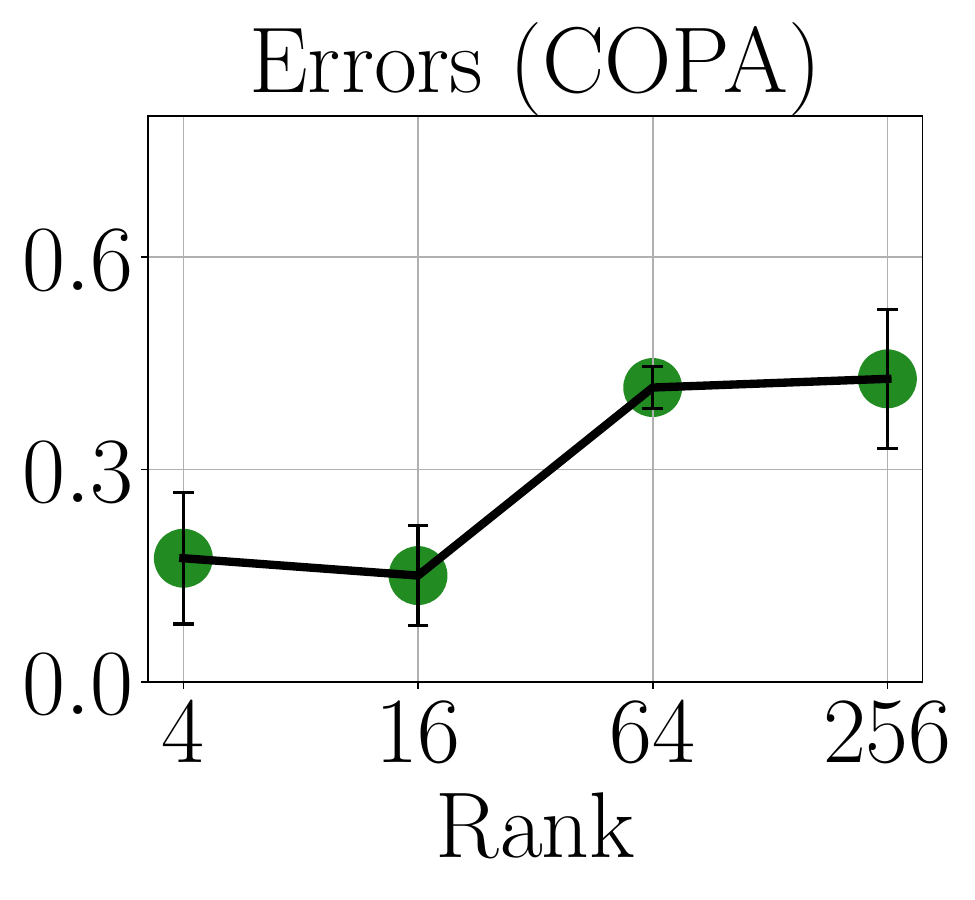}
    \end{subfigure}\vfill
    \begin{subfigure}[b]{\textwidth}
    \includegraphics[width=\textwidth]{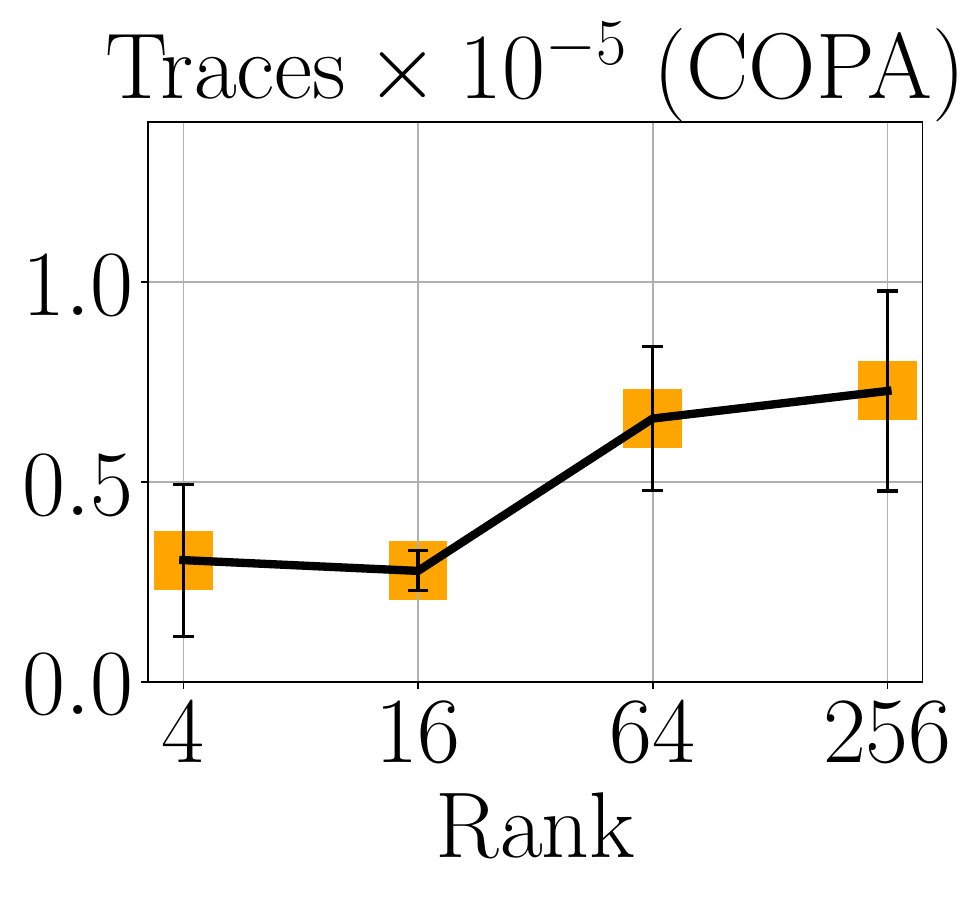}
    \end{subfigure}
    \end{subfigure}
    \begin{subfigure}[b]{0.155\textwidth}
    \centering
    \begin{subfigure}[b]{\textwidth}
    \includegraphics[width=\textwidth]{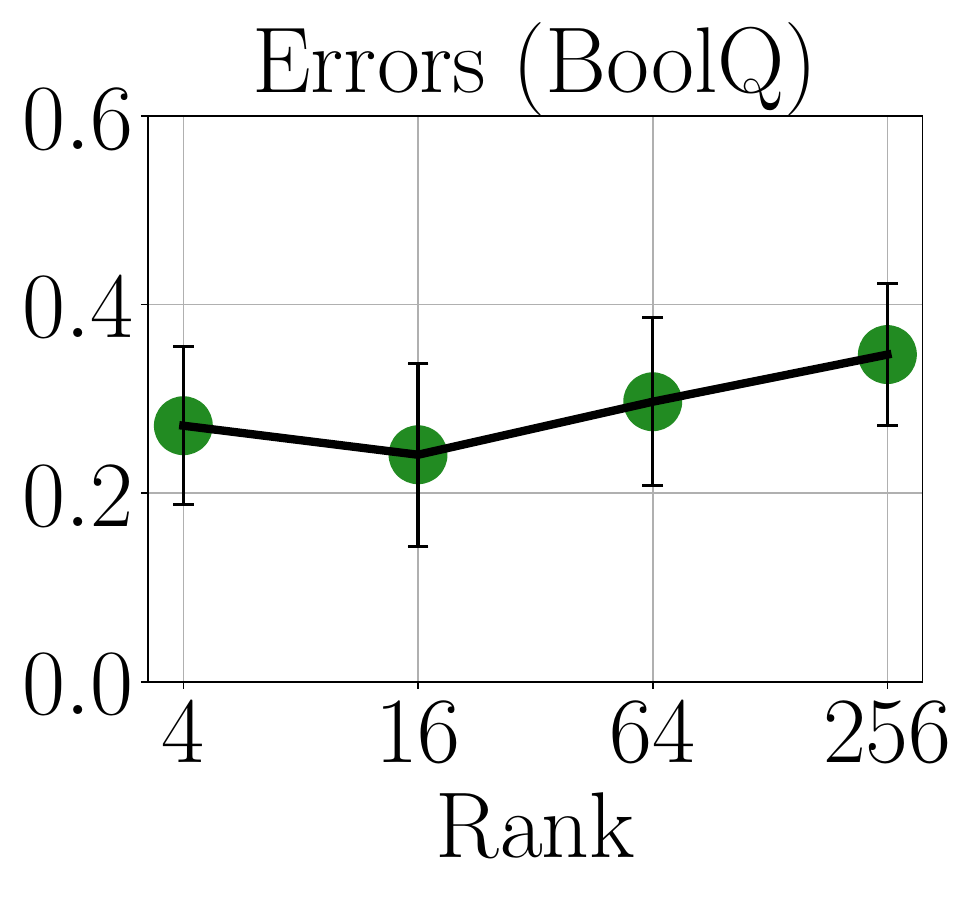}
    \end{subfigure}\vfill
    \begin{subfigure}[b]{\textwidth}
    \includegraphics[width=\textwidth]{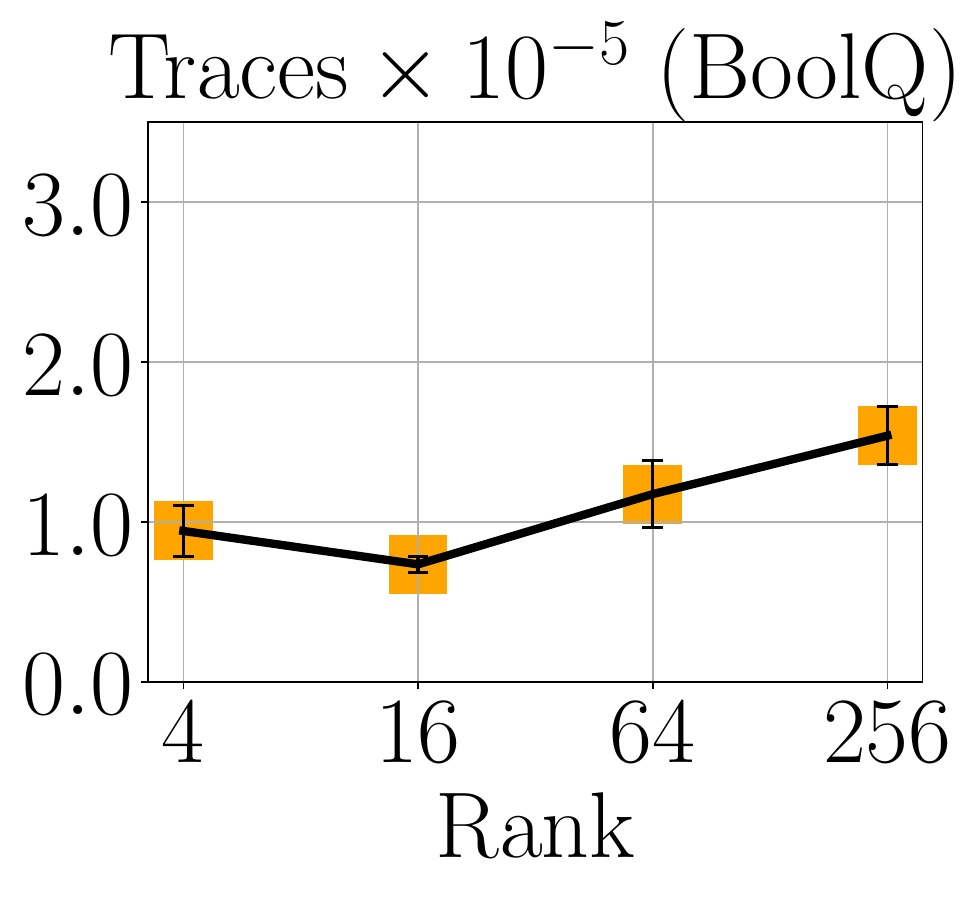}
    \end{subfigure}
    \end{subfigure}
    \caption{We show the empirical generalization errors (termed Error in the figures) and estimated Hessian traces of the loss for fine-tuned adapters (using QAdapter on Llama-3-8B) across various hidden dimensions. Smaller adapters achieve the lowest generalization errors. Additionally, the Hessian trace values correlate with generalization errors, suggesting that smaller adapters tend to converge to flatter minima.}\label{fig_generalization_errors_v3}
\end{figure}

We observe similar evaluation of generalization errors and sharpness measures on the BoolQ dataset. %
We also notice consistent results of using QAdapter, which are reported in Figure \ref{fig_generalization_errors_v3}. 

\subsection{Measuring transfer effects}

Consider fine-tuning a foundation model to predict $n$ individual tasks.
Given a fine-tuning protocol such as QLoRA, denoted by $f$, and a subset of tasks $S \subseteq \set{1, 2, \dots, n}$, let $f_i(S)$ represent the loss value of the fine-tuned model evaluated on task $i$, for any $i \in S$. 
We define the {positive transfer rate} of $f$ as follows.
First, we say that tasks $S$ provide a positive transfer to task $i$ if $f_i(S)$ is lower than the single-task loss value of $f_i{\set{i}}$. Otherwise, we say that the transfer is negative.
Then, imagine a scenario where we examine a list of task subsets $S_1, \ldots, S_k$ (for instance, in the simplest case, $m$ could be $1$ and $S_1$ could include all the tasks as $S_1 = \set{1, 2, \dots, n}$).
With this notion of positive/negative transfer, we evaluate the positive transfer rate averaged over the $k$ subsets as:
\[ \frac{1}{k} \sum_{j=1}^k \frac{\big|\set{i \in S_j: f_i(S_j) < f_i(\set{i})  }\big|}{\big|S_j\big|}. \]
With this definition in hand, next, we will examine the transfer rate of various fine-tuning methods, along with their memory usage. 
In a nutshell, we observe that there is an intricate trade-off between transfer and memory, as we will show below.
Our overall goal is to design a boosting system that optimizes positive transfer on a base fine-tuning protocol with minimal computation and memory overhead.

\paragraph{Computational cost vs. transfer.}
We now report the memory usage of various fine-tuning algorithms. We evaluate the memory usage of an ensemble of adapters, which involves using multiple adapters and combining their outputs through weighted averaging. 
Our findings using Llama-3-8B as the base model (the results obtained with other base models are qualitatively similar) are the following. 
As expected, with full fine-tuning, memory size scales as $21.5$ GB (size of Llama-8B) times $m$.
With PEFT, this reduces to $21.5 + 2.8 (m - 1)$ GB for LoRA and $21.5 + 6.5 (m - 1)$ for adapter tuning.
Applying 4-bit quantization to the base model further reduces the memory by 53\% using LoRA and 56\% using adapter tuning.

Next, we report the transfer rates from the above fine-tuning procedures. %
Recall that this depends on the specification of the subsets.
We first consider $k=1$ and $S_1 = \set{1, 2, \dots, n}$.
Interestingly, we note that the transfer rate decreases with (quantized) PEFT compared to full fine-tuning.
For full fine-tuning, the positive transfer rate is $6/10=60\%$. With LoRA/adapter, this decreases to $40\%$.
For QLoRA and QAdapter, this further decreases to $30\%$ and $10\%$.

Next, we consider $k=45$ and let the list of subsets be $\set{1, 2}, \set{1, 3}, \dots, \set{n-1, n}$, which includes all pairwise combinations. %
We find that for full fine-tuning, the positive transfer rate is 56\%. For LoRA/adapter, this decreases to 38\% and 48\%, and 26{\%} and 22{\%} with quantization.
In addition, we consider $f$ by averaging the weights of two adapters, each trained on a single task.
This further reduces positive transfer across tasks.

\begin{figure}[t!]
    \centering
    \begin{subfigure}[b]{0.485\textwidth}
    \begin{subfigure}[b]{0.48\textwidth}
    \centering    
    \includegraphics[width=0.80\textwidth]{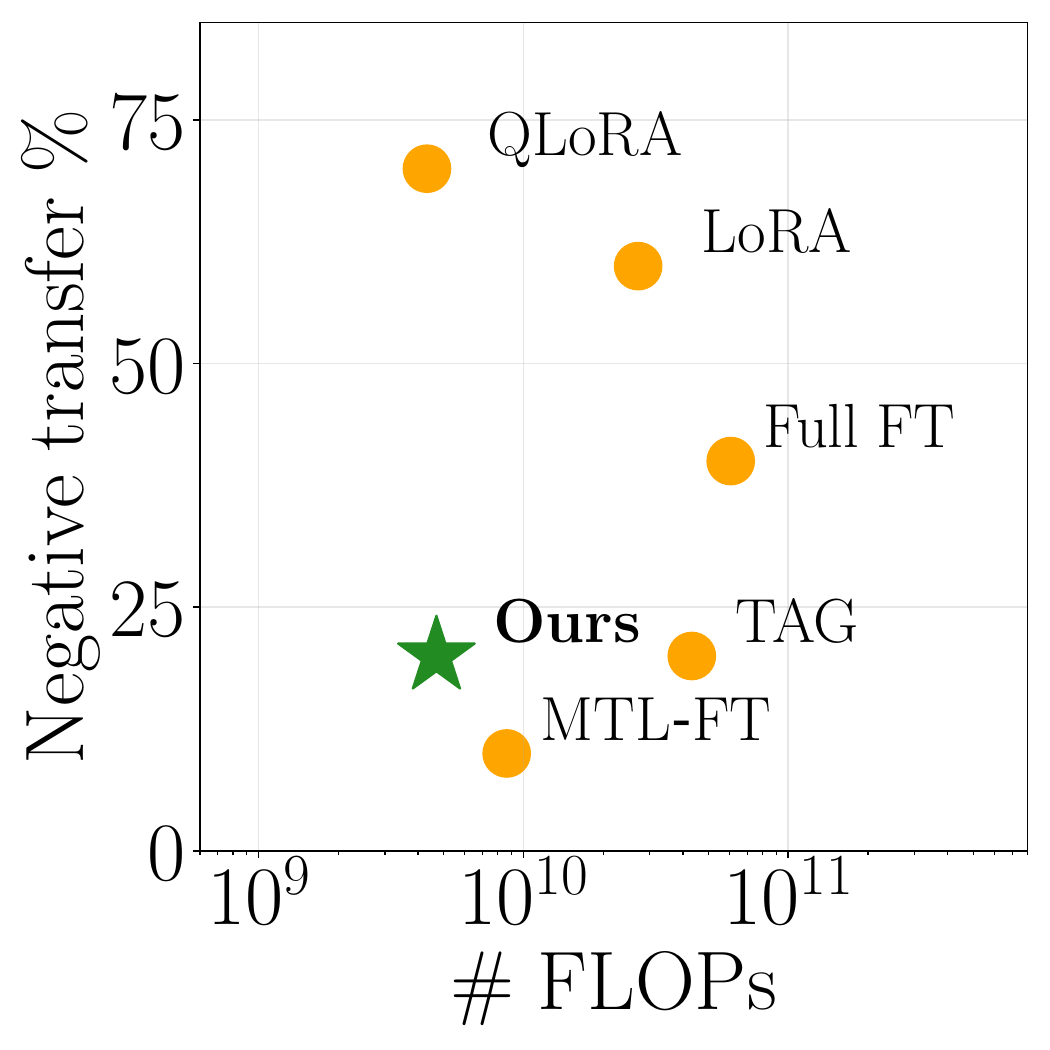}
    \end{subfigure}\hfill
    \begin{subfigure}[b]{0.48\textwidth}
    \centering    
    \includegraphics[width=0.80\textwidth]{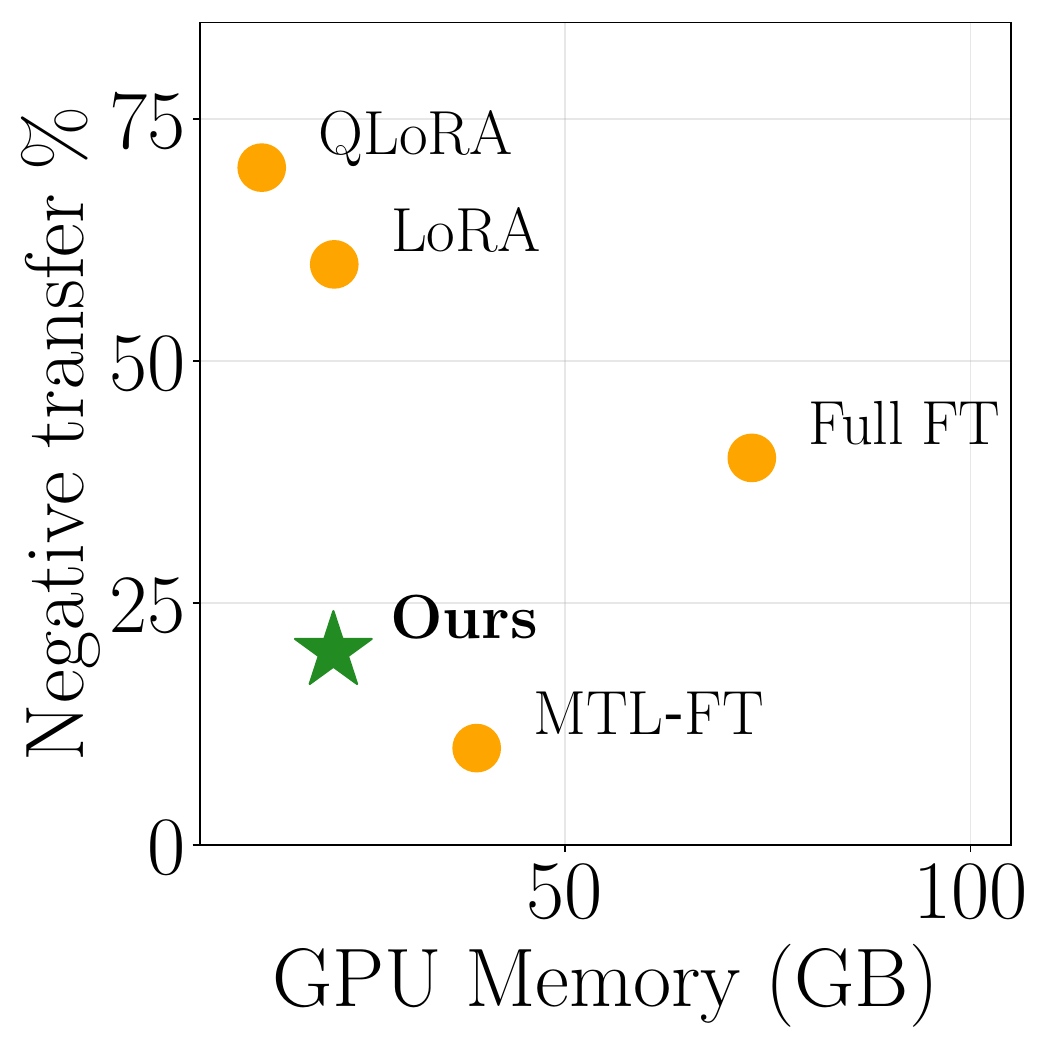}
    \end{subfigure}
    \subcaption{Using QLoRA with base protocol}\label{alg_comparison_computation_memory_transfer_qlora_2} %
    \end{subfigure}
    \hfill
    \begin{subfigure}[b]{0.485\textwidth}
    \begin{subfigure}[b]{0.48\textwidth}
    \centering    
    \includegraphics[width=0.80\textwidth]{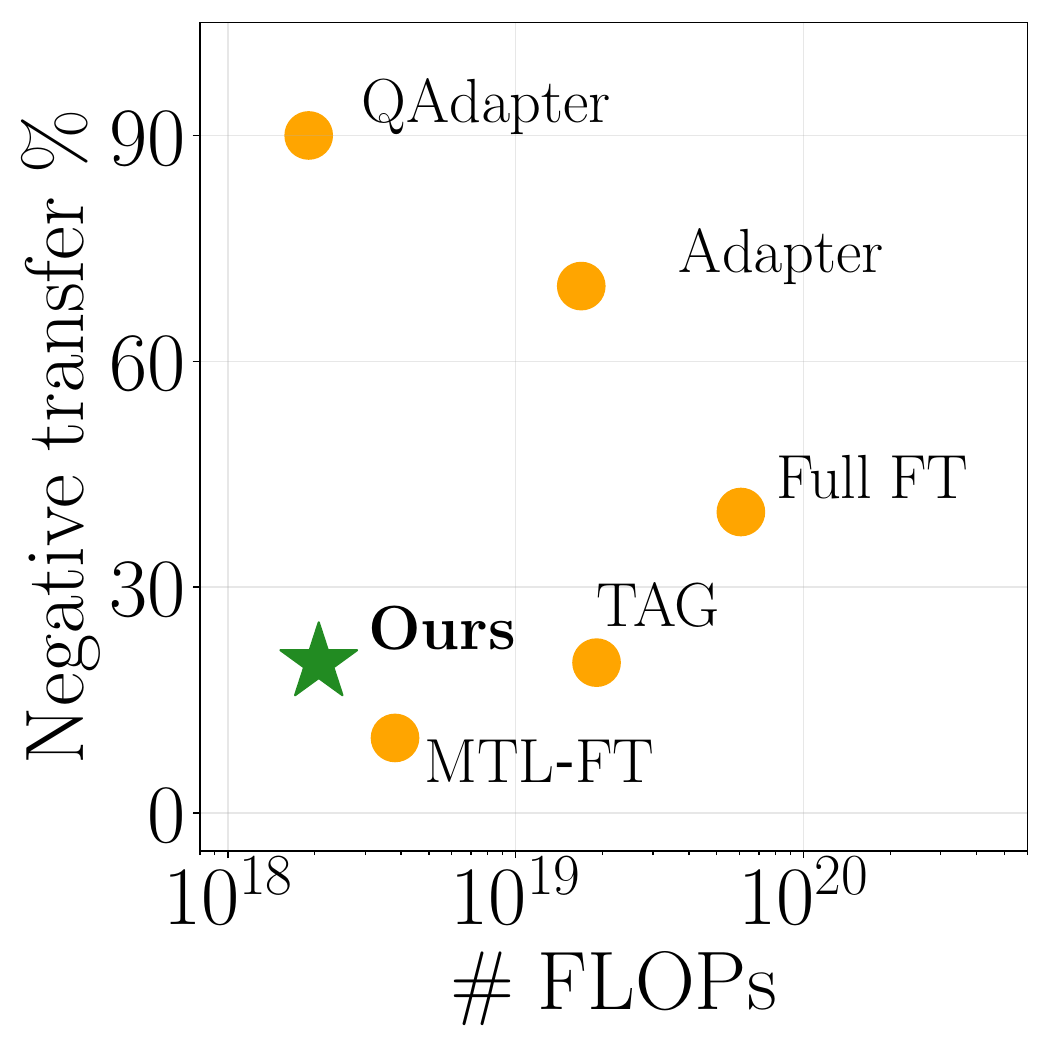}
    \end{subfigure}\hfill
    \begin{subfigure}[b]{0.48\textwidth}
    \centering    
    \includegraphics[width=0.80\textwidth]{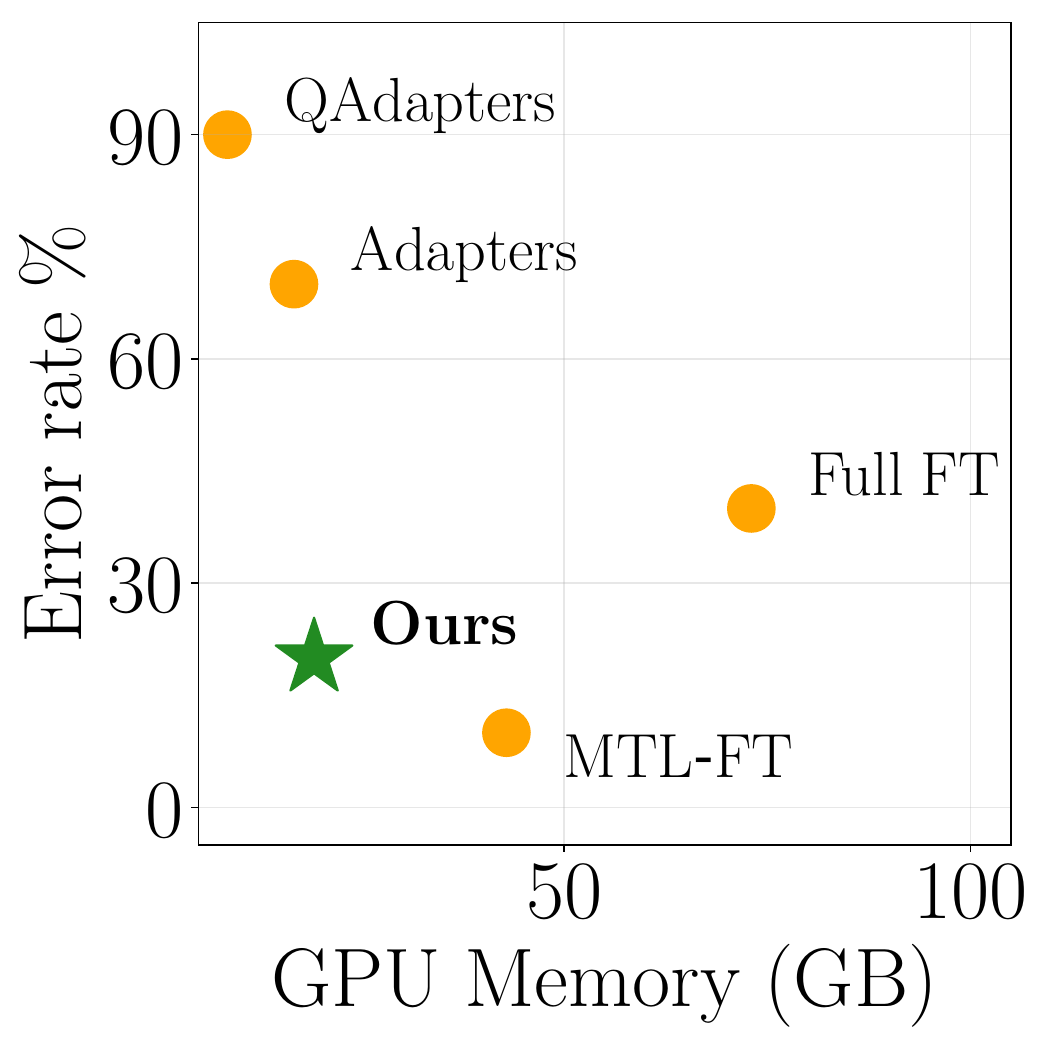}
    \end{subfigure}
    \subcaption{Using QAdapter as base protocol}\label{alg_comparison_computation_memory_transfer_qadapter_2}
    \end{subfigure}
    \caption{We illustrate the trade-off between transfer, computation cost, and GPU memory between our approach and baselines. 
    \algname~achieves the best trade-off between task transfers and computation/memory costs. 
    This experiment uses Llama-3-8B as the base model.}\label{fig_tradeoff_positive_transfer}
\end{figure}

Our approach also achieves the best trade-off between the positive transfer rate and computation/memory cost, illustrated in Figure \ref{fig_tradeoff_positive_transfer}. 
Compared to QLoRA and QAdapter, our approach consistently boosts the positive transfer rates from 30\% to 80\%. Our approach achieves a comparable positive transfer rate as MTL-FT (90\%) and also improves over the positive transfer rate of full model fine-tuning by 60\%.

\begin{figure}[h!]
    \centering
    \begin{subfigure}[b]{0.235\textwidth}
    \centering    
    \includegraphics[width=0.99\textwidth]{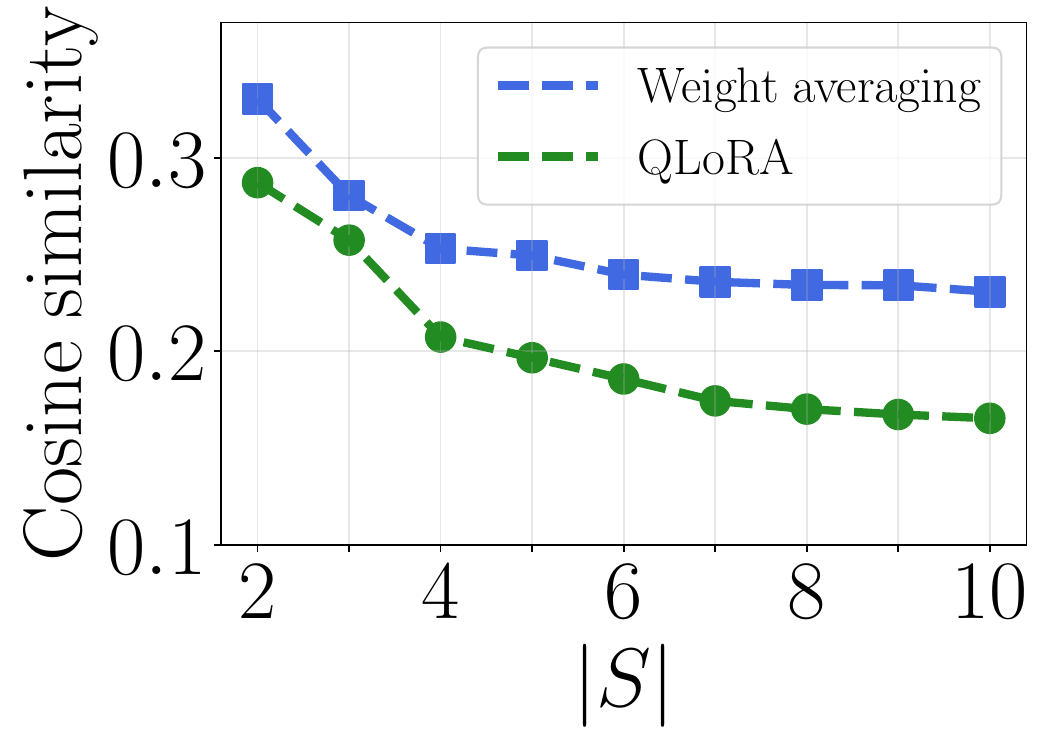}
    \end{subfigure}\hfill
    \begin{subfigure}[b]{0.235\textwidth}
    \centering    
    \includegraphics[width=0.99\textwidth]{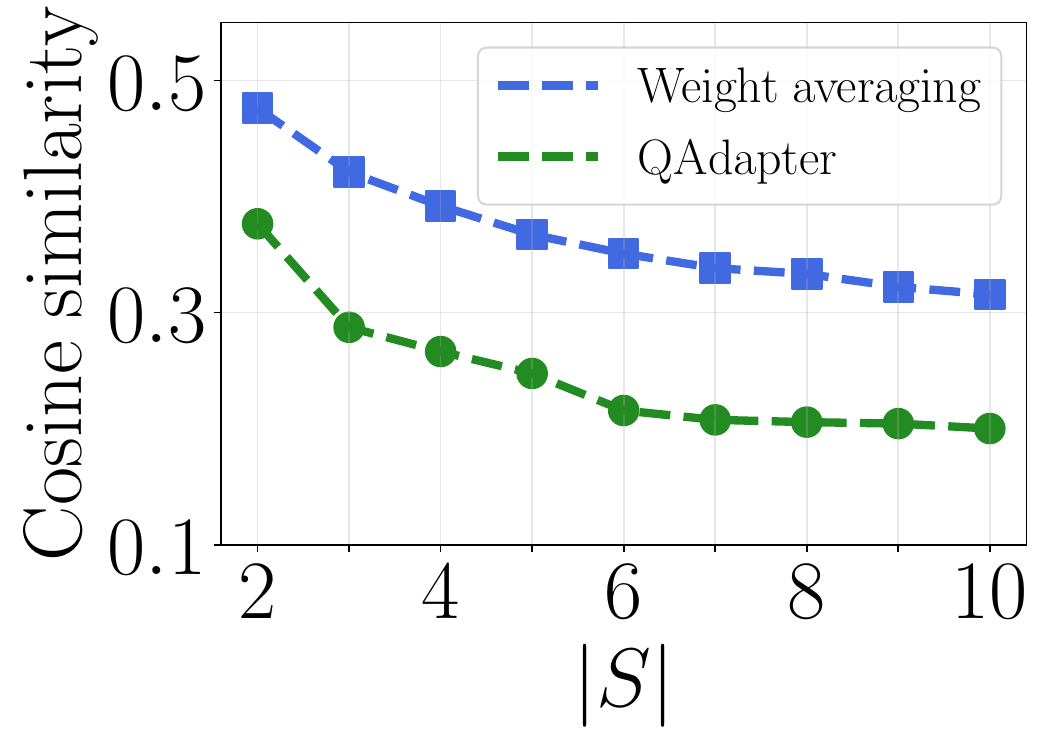}
    \end{subfigure}
    \caption{We measure the average similarity between models fine-tuned on random subsets $S$ and single-task models. We observe that the similarity score becomes lower as the size of $S$ increases.}
    \label{fig_weight_conflicts}
\end{figure}

Next, we report the cosine similarity score between adapter weights to explain the decrease in transfer rates. We use the base model Llama-3-1B and evaluate fine-tuned QLoRA and QAdapter weights.
Given two weight matrices from two models, we compute the score between their rank-$r$ decompositions. For QLoRA, we apply the decomposition to the weight matrix, $\Delta W = BA$, with the same rank as $A$, and average the scores over layers.
We measure the similarity between models fine-tuned on random subsets $S$ and the single-task fine-tuned models, varying the size of $S$ from 2 to 10. 
We also compute the scores using weight averaging as $f$.
As illustrated in Figure \ref{fig_weight_conflicts}, for both fine-tuning and weight averaging, the similarity score becomes lower as the size of $S$ increases. 

{}

\end{document}